\documentclass[journal,transmag]{IEEEtran}
\usepackage{comment}
\usepackage{multirow}
\usepackage{amsmath}
\usepackage{wrapfig}
\usepackage{array}
\newcolumntype{?}{!{\vrule width 1pt}}
\usepackage{wasysym}
\usepackage{makecell}
\usepackage{arydshln}
\usepackage{color}

\usepackage{graphicx}

\newcommand{\mcrot}[4]{\multicolumn{#1}{#2}{\rlap{\rotatebox{#3}{#4}~}}} 
\newcommand*{\twoelementtable}[3][l]%
{%
    \begin{tabular}[t]{@{}#1@{}}%
        #2\tabularnewline
        #3%
    \end{tabular}%
}
\newcommand{\specialcell}[2][c]{%
  \begin{tabular}[#1]{@{}c@{}}#2\end{tabular}}

\usepackage{pstricks}
\usepackage{pst-plot}

\newcommand{\moon}[1]{%
\begingroup%
\psset{unit=1ex}%
\psset{linewidth=0.02em}%
\def\synodicmonth{29.530588853}%
\def\lunarage{#1}%
\edef\waxswitch{\ifdim\lunarage pt>.5\dimexpr\synodicmonth pt 1\else -1\fi}%
\begin{pspicture}[showgrid=false](-1.1,-.65)(1.1,1)%
    \parametricplot[plotpoints=91]{0}{360}{t sin t cos}%
    \pscustom[fillstyle=solid,fillcolor=black]{%
    \parametricplot[plotpoints=91]{\waxswitch\space 180 mul}{0}{t sin t cos}%
    \parametricplot[plotpoints=91]{0}{180}{t sin \lunarage\space \synodicmonth\space div 1 \waxswitch\space add 4 div add 360 mul cos mul t cos}%
    }
\end{pspicture}%
\endgroup%
}

\begin{document}
\title{Deep Learning for Video Game Playing}
\author{Niels Justesen$^{1}$, Philip Bontrager$^{2}$, Julian Togelius$^{2}$, Sebastian Risi$^{1}$\\
$^{1}$IT University of Copenhagen, Copenhagen\\
$^{2}$New York University, New York\vspace{-0.2in}
}

\maketitle

\begin{abstract}
In this article, we review recent Deep Learning advances in the context of how they have been applied to play different types of video games such as first-person shooters, arcade games, and real-time strategy games. We analyze the unique requirements that different game genres pose to a deep learning system and highlight important open challenges in the context of applying these machine learning methods to video games, such as general game playing, dealing with extremely large decision spaces and sparse rewards. 
\end{abstract}

\vspace{-1.5em}

\section{Introduction}
Applying AI techniques to games is now an established research field with multiple conferences and dedicated journals. In this article, we review recent advances in deep learning for video game playing and employed game research platforms while highlighting important open challenges. Our motivation for writing this article is to review the field from the perspective of different types of games, the challenges they pose for deep learning, and how deep learning can be used to play these games. A variety of review articles on deep learning exists~\cite{goodfellow2016deep,lecun2015deep,schmidhuber2015deep}, as well as surveys on reinforcement learning~\cite{sutton1998reinforcement} and deep reinforcement learning~\cite{li2017deep}, here we focus on these techniques applied to video game playing.


In particular, in this article, we focus on game problems and environments that have been used extensively for DL-based Game AI, such as Atari/ALE, Doom, Minecraft, StarCraft, and car racing. Additionally, we review existing work and point out important challenges that remain to be solved. We are interested in  approaches that aim to play a particular \emph{video game} well (in contrast to board games such as Go, etc.), from pixels or feature vectors, without an existing forward model. Several game genres are analyzed to point out the many and diverse challenges they pose to human and machine players.

It is important to note that there are many uses of AI in and for games that are not covered in this article; Game AI is a large and diverse field~\cite{yannakakis2017artificial,yannakakis2015panorama,miikkulainen2006computational,galway2008machine,munoz2013learning}. 
This article is focused on deep learning methods for playing video games well, while there is plenty of research on playing games in a believable, entertaining or human-like manner~\cite{hingston2012believable}. AI is also used for modeling players' behavior, experience or preferences~\cite{yannakakis2013player}, or generating game content such as levels, textures or rules~\cite{shaker2016procedural}. Deep learning is far from the only AI method used in games. Other prominent methods include Monte Carlo Tree Search~\cite{browne2012survey} and evolutionary computation~\cite{risi2015neuroevolution,lucas2006evolutionary}. In what follows, it is important to be aware of the limitations of the scope of this article.

The paper is structured as follows: The next section gives an overview of different deep learning methods applied to games, followed by the different research platforms that are currently in use. Section~\ref{sec:game_playing} reviews the use of DL methods in different video game types and Section~\ref{sec:history} gives a historical overview of the field. We conclude the paper by pointing out important open challenges in Section~\ref{sec:open} and a conclusion in Section~\ref{sec:conclusion}.





\section{Deep Learning in Games Overview}
This section gives a brief overview of neural networks and machine learning in the context of games. First, we describe common neural network architectures followed by an overview of the three main categories of machine learning tasks: supervised learning, unsupervised learning, and reinforcement learning. Approaches in these categories are typically based on gradient-descent optimization. We also highlight evolutionary approaches as well as a few examples of hybrid approaches that combine several optimization techniques.

\subsection{Neural Network Models}

Artificial neural networks (ANNs) are general purpose functions that are defined by their network structure and the weight of each graph edge. Because of their generality and ability to approximate any continuous real-valued function (given enough parameters), they have been applied to a variety of tasks, including video game playing. 
%
The architectures of these ANNs can roughly be divided into two major categories: feedforward and recurrent neural networks (RNN). Feedforward networks take a single input, for example, a representation of the game state, and outputs probabilities or values for each possible action. 
Convolutional neural networks (CNN) consists of trainable filters and is suitable for processing image data such as pixels from a video game screen.


RNNs are typically applied to time series data, in which the output of the network can depend on the network's activation from previous time-steps \cite{williams1989learning, lecun1989backpropagation}. The training process is similar to feedforward networks, except that the network's previous hidden state is fed back into the network together with the next input. This allows the network to become context-aware by memorizing the previous activations, which is useful when a single observation from a game does not represent the complete game state. For video game playing, it is common to use a stack of convolutional layers followed by recurrent layers and fully-connected feed-forward layers.

The following sections will give a brief overview of different optimization methods which are commonly used for learning game-playing behaviors with deep neural networks. These methods search for the optimal set of parameters to solve some problem. Optimization can also be used to find hyper-parameters, such as network architecture and learning parameters, and is well studied within deep learning \cite{bergstra2011algorithms, bergstra2013making}.

\subsection{Optimizing Neural Networks}

\subsubsection{Supervised Learning}

In supervised learning a model is trained from examples. During training, the model is asked to make a decision for which the correct answer is known. The error, i.e. difference between the provided answer and the ground truth, is used as a loss to update the model. The goal is to achieve a model that can generalize beyond the training data and thus perform well on examples it has never seen before. Large data sets usually improve the model's ability to generalize. 

In games, this data can come from play traces  \cite{apprenticeshipBogdanovic2015} (i.e.\ humans playing through the game while being recorded), allowing the agent to learn the mapping from the input state to output actions based on what actions the human performed in a given state. If the game is already solved by another algorithm, it can be used to generate training data, which is useful if the first algorithm is too slow to run in real-time. 
While learning to play from existing data allows agents to quickly learn best practices, it is often brittle; the data available can be expensive to produce and may be missing key scenarios the agent should be able to deal with. For gameplay, the algorithm is limited to the strategies available in the data and cannot explore new ones itself. Therefore, in games, supervised algorithms are often combined with additional training through reinforcement learning algorithms \cite{silver2016mastering}.

Another application of supervised learning in games is to learn the state transitions of a game. Instead of providing the action for a given state, the neural network can learn to predict the next state for an action-state pair. Thus, the network is essentially learning a model of the game, which can then be used to play the game better or to perform planning \cite{ha2018world}.


\subsubsection{Unsupervised Learning}
Instead of learning a mapping between data and its labels, the objective in unsupervised learning is to discover patterns in the data. 
These algorithms can learn the distribution of features for a dataset, which can be used to cluster similar data, compress data into its essential features, or create new synthetic data that is characteristic of the original data. For games with sparse rewards (such as Montezuma's Revenge), learning from data in an unsupervised fashion is a potential solution and an important open deep learning challenge.

A prominent unsupervised learning technique in deep learning is the \emph{autoencoder}, which is a neural network that attempts to learn the identity function such that the output is identical to the input \cite{le1987modeles, rumelhart1986general}. The network consists of two parts: an encoder that maps the input $x$ to a low-dimensional hidden vector $h$, and a decoder that attempts to re-construct $x$ from $h$. 
The main idea is that by keeping $h$ small, the network has to learn to compress the data and therefore learn a good representation. Researchers are beginning to apply such unsupervised algorithms to games to help distill high dimensional data to more meaningful lower dimensional data, but this research direction is still in its early stages \cite{ha2018world}. For a more detailed overview of supervised and unsupervised learning see \cite{schmidhuber2015deep,goodfellow2016deep}.


\subsubsection{Reinforcement Learning Approaches}


In reinforcement learning (RL) an agent learns a behavior by interacting with an environment that provides a reward signal back to the agent. A video game can easily be modeled as an environment in an RL setting, wherein players are modeled as agents with a finite set of actions that can be taken at each step and the reward signal can be determined by the game score.

\begin{figure}
\begin{center}
  \includegraphics[width=.9\columnwidth]{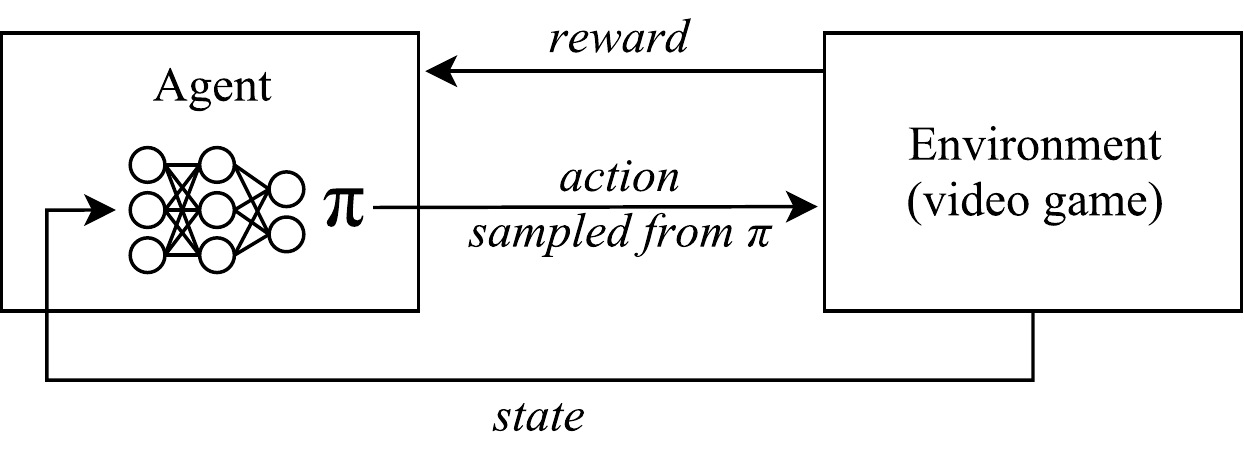} 
  \vspace{-1.0em}
  \caption{The reinforcement learning framework where an agent's policy $\pi$ is determined by a deep neural network. The state of the environment, or an observation such as screen pixels, is fed as input to the agent's policy network. An action is sampled from the policy network's output $\pi$ where after it receives a reward and the subsequent game state. The goal is to maximize the cumulated rewards. The reinforcement learning algorithm updates the policy (network parameters) based on the reward.} 
  \label{fig:rl}
  \vspace{-2.0em}
\end{center}
\end{figure}

In RL, the agent relies on the reward signal. These signals can occur frequently, such as the change in score within a game, or it can occur infrequently, such as whether an agent has won or lost a game. Video games and RL go well together since most games give rewards for successful strategies.  
Open world games do not always have a clear reward model and are thus challenging for RL algorithms.

A key challenge in applying RL to games with sparse rewards is to determine how to assign credit to the many previous actions when a reward signal is obtained. The reward $R(s)$ for state $s$, needs to be propagated back to the actions that lead to the reward. Historically, there are several different ways this problem is approached which are described below. If an environment can be described as a Markov Decision Process (MDP), then the agent can build a probability tree of future states and their rewards. The probability tree can then be used to calculate the utility of the current state. For an RL agent this
means learning the model $P(s' \vert s, a)$, where $P$ is the probability of state $s'$ given state $s$ and action $a$. With a model $P$, utilities can be calculated by 
$$U(s) = R(s) + \gamma \max_a \sum_{s'} P(s' \vert s, a)U(s'),$$
where $\gamma$ is the discount factor for the utility of future states. This algorithm, known as Adaptive Dynamic Programming, can converge rather quickly as it directly handles the credit assignment problem \cite{sutton1998reinforcement}. The issue is that it has to build a probability tree over the whole problem space and is therefore intractable for large problems. As the games covered in this work are considered "large problems", we will not go into further detail on this algorithm.

Another approach to this problem is temporal difference (TD) learning. In TD learning, the agent learns the utilities \textit{U} directly based off of the observation that the current utility is equal to the current reward plus the utility value of the next state \cite{sutton1998reinforcement}. Instead of learning the state transition model $P$, it learns to model the utility $U$, for every state. The update equation for $U$ is:
$$U(s) = U(s) + \alpha(R(s) + \gamma U(s') - U(s)),$$ 
where $\alpha$ is the learning rate of the algorithm. The equation above does not take into account how $s'$ was chosen. If a reward is found at $s_t$, it will only affect $U(s_t)$. The next time the agent is at $s_{t-1}$, then $U(s_{t-1})$ will be aware of the future reward. This will propagate backward over time. Likewise, less common transitions will have less of an impact on utility values. Therefore, $U$ will converge to the same values as are obtained from ADP, albeit slower.

There are alternative implementations of TD that learn rewards for state-action pairs. This allows an agent to choose an action, given the state, with no model of how to transition to future states. For this reason, these approaches are referred to as model-free methods. A popular model-free RL method is Q-learning \cite{watkins1992q} where the utility of a state is equal to the maximum Q-value for a state. The update equation for Q-learning is: 
$$Q(s, a) = Q(s, a) + \alpha(R(s) + \gamma \max_{a'} Q(s',a') - Q(s, a)).$$
In Q-learning, the future reward is accounted for by selecting the best-known future state-action pair. In a similar algorithm called SARSA (State-Action-Reward-State-Action), $Q(s,a)$ is updated only when the next $a$ has been selected and the next $s$ is known \cite{rummery1994line}. This action pair is used instead of the maximum Q-value. This makes SARSA an \emph{on-policy} method in contrast to Q-learning which is \emph{off-policy} because SARSA's Q-value accounts  for the agent's own policy. 

Q-learning and SARSA can use a neural network as a function approximator for the Q-function. The given Q update equation can be used to provide the new "expected" Q value for a state-action pair. The network can then be updated as it is in supervised learning.

An agent's policy $\pi(s)$ determines which action to take given a state $s$. For Q-learning, a simple policy would be to always take the action with the highest Q-value. Yet, early on in training, Q-values are not very accurate and an agent could get stuck always exploiting a small reward. A learning agent should prioritize exploration of new actions as well as the exploitation of what it has learned. This problem is known as a multi-armed bandit problem and has been well explored. The $\epsilon$-greedy strategy is a simple approach that selects the (estimated) optimal action with $\epsilon$ probability and otherwise selects a random action. 

One approach to RL is to perform gradient descent in the policy's parameter space. Let $\pi_\theta(s, a)$ be the probability that action $a$ is taken at state $s$ given parameters $\theta$. 
The basic policy gradient algorithm from the REINFORCE family of algorithms \cite{reinforceWilliams1992} updates $\theta$ using the gradient $\nabla_\theta\sum_a \pi_\theta(s, a)R(s)$ where $R(s)$ is the discounted cumulative reward obtained from $s$ and forward. In practice, a sample of possible actions from the  policy is taken and it is updated to increase the likelihood that the more successful actions are returned in the future. This lends itself well to neural networks as $\pi$ can be a neural network and $\theta$ the network weights.

Actor-Critic methods combine the policy gradient approach with TD learning, where an actor learns a policy $\pi_\theta(s,a)$ using the policy gradient algorithm, and the critic learns to approximate $R$ using TD-learning \cite{actorSutton1999}. Together, they are an effective approach to iteratively learning a policy. In actor-critic methods there can either be a single network to predict both $\pi$ and $R$, or two separate networks. For an overview of reinforcement learning applied to deep neural networks we suggest the article by Arulkumaran et al.~\cite{arulkumaran2017brief}.

\subsubsection{Evolutionary Approaches}
The optimization techniques discussed so far rely on gradient descent, based on differentiation of a defined error. However, derivative-free optimization methods such as evolutionary algorithms have also been widely used to train neural networks, including, but not limited to, reinforcement learning tasks. This approach, often referred to as neuroevolution (NE), can optimize a network's weights as well as their topology/architecture. Because of their generality, NE approaches have been applied extensively to different types of video games. For a complete overview of this field, we refer the interested reader to our NE survey paper~\cite{risi2015neuroevolution}.

Compared to gradient-descent based training methods, NE approaches have the benefit of not requiring the network to be differentiable and can be applied to both supervised, unsupervised and reinforcement learning problems. 
The ability to evolve the topology, as well as the weights, potentially offers a way of automating the development of neural network architecture, which currently requires considerable domain knowledge. The 
promise of these techniques is that evolution could find a neural network topology that is better at playing a certain game than existing human-designed architectures.
While NE has been traditionally applied to problems with lower input dimensionality than typical deep learning approaches, recently Salimans et al.~\cite{salimans2017evolution} showed that evolution strategies, which rely on parameter-exploration through stochastic noise instead of calculating gradients, can achieve results competitive to current deep RL approaches for Atari video game playing, given enough computational resources. 

\subsubsection{Hybrid Learning Approaches}
More recently researchers have started to investigate hybrid approaches for video game playing, which combine deep learning methods with other machine learning approaches. 
%
Both Alvernaz and Togelius~\cite{alvernaz2017autoencoder} and Poulsen et al.~\cite{Poulsen2017} experimented with combining a deep network trained through gradient descent feeding a condensed feature representation into a network trained through artificial evolution. These hybrids aim to combine the best of both approaches as deep learning methods are able to learn directly from high-dimensional input, while evolutionary methods do not rely on differentiable architectures and work well in games with sparse rewards. Some results suggest that gradient-free methods seem to be better in the early stages of training to avoid premature convergence while gradient-based methods may be better in the end when less exploration is needed \cite{such2017deep}.

Another hybrid method for board game playing was AlphaGo \cite{silver2016mastering} that relied on deep neural networks and tree search methods to defeat the world champion in Go, and \cite{fragkiadaki2015learning} that applies planning on top of a predictive model. 

In general, the hybridization of \emph{ontogenetic} RL (such as Q-learning) with \emph{phylogenetic} methods (such as evolutionary algorithms) has the potential to be very impactful as it could enable concurrent learning on different timescales~\cite{togelius2009ontogenetic}.

\section{Game Genres and Research Platforms}

The fast progression of deep learning methods is undoubtedly due to the convention of comparing results on publicly available datasets. A similar convention in game AI is to use game environments to compare game playing algorithms, in which methods are ranked based on their ability to score points or win in games. Conferences like the IEEE Conference on Computational Intelligence and Games run popular competitions in a variety of game environments. 

\begin{figure*}
\begin{center}
  \includegraphics[width=\textwidth]{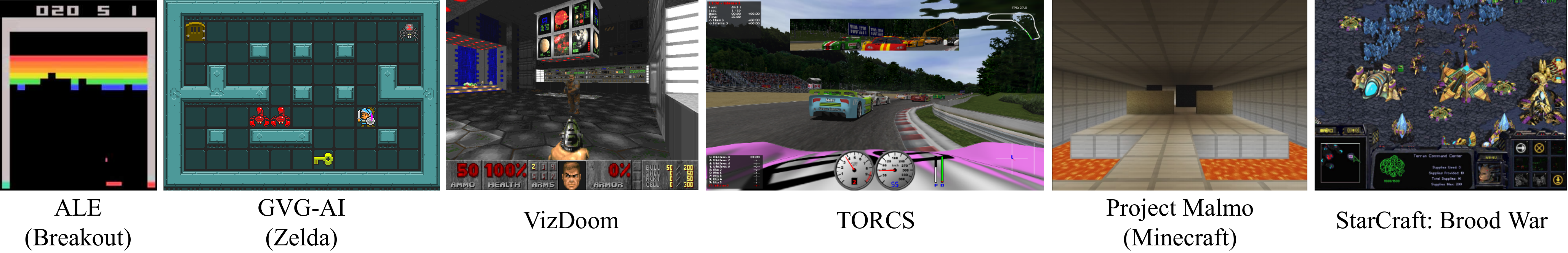} 
  \caption{Screenshots of selected games and frameworks used as research platforms for research in deep learning.} 
  \label{fig:games}
  \vspace{-2.0em}
\end{center}
\end{figure*}

This section describes popular game genres and research platforms, used in the literature, that are relevant to deep learning; some examples are shown in Figure~\ref{fig:games}. For each genre, we briefly outline what characterizes that genre and describe the challenges faced by algorithms playing games of the genre. The video games that are discussed in this paper have to a large extent supplanted an earlier generation of simpler control problems that long served as the main reinforcement learning benchmarks but are generally too simple for modern RL methods.
In such classic control problems, the input is a simple feature vector, describing the position, velocity, and angles etc. Popular platforms for such problems are rllab \cite{duan2016benchmarking}, which includes classic problems such as pole balancing and the mountain car problem, and MuJoCo (Multi-Joint dynamics with Contact), a physics engine for complex control tasks such as the humanoid walking task \cite{todorov2012mujoco}. 

\subsection{Arcade Games}

Classic arcade games, of the type found in the late seventies' and early eighties' arcade cabinets, home video game consoles and home computers, have been commonly used as AI benchmarks within the last decade. Representative platforms for this game type are the Atari 2600, Nintendo NES, Commodore 64 and ZX Spectrum. Most classic arcade games are characterized by movement in a two-dimensional space (sometimes represented isometrically to provide the illusion of three-dimensional movement), heavy use of graphical logics (where game rules are triggered by the intersection of sprites or images), continuous-time progression, and either continuous-space or discrete-space movement. The challenges of playing such games vary by game. Most games require fast reactions and precise timing, and a few games, in particular, early sports games such as \emph{Track \& Field} (Konami, 1983) rely almost exclusively on speed and reactions. Many games require prioritization of several co-occurring events, which requires some ability to predict the behavior or trajectory of other entities in the game. This challenge is explicit in e.g. \emph{Tapper} (Bally Midway, 1983) but also in different ways part of platform games such as \emph{Super Mario Bros} (Nintendo, 1985) and shooters such as \emph{Missile Command} (Atari Inc., 1980). Another common requirement is navigating mazes or other complex environments, as exemplified clearly by games such as \emph{Pac-Man} (Namco, 1980) and \emph{Boulder Dash} (First Star Software, 1984). Some games, such as \emph{Montezuma's Revenge} (Parker Brothers, 1984), require long-term planning involving the memorization of temporarily unobservable game states. Some games feature incomplete information and stochasticity, others are completely deterministic and fully observable.

The most notable game platform used for deep learning methods is the Arcade Learning Environment (ALE) \cite{bellemare2013arcade}. ALE is built on top of the Atari 2600 emulator Stella and contains more than 50 original Atari 2600 games. The framework extracts the game score, 160$\times$210 screen pixels and the RAM content that can be used as input for game playing agents. ALE was the main environment explored in the first deep RL papers that used raw pixels as input. By enabling agents to learn from visual input, ALE thus differs from classic control problems in the reinforcement learning literature, such as the Cart Pole and Mountain Car problems. An overview and discussion of the ALE environment can be found in \cite{machado2017revisiting}.  
 
Another platform for classic arcade games is the Retro Learning Environment (RLE) that currently contains seven games released for the Super Nintendo Entertainment System (SNES) \cite{bhonker2016playing}. Many of these games have 3D graphics and the controller allows for over 720 action combinations. SNES games are thus more complex and realistic than Atari 2600 games but RLE has not been as popular as ALE. 


The General Video Game AI (GVG-AI) framework \cite{torrado2018deep} allows for easy creation and modification of games and levels using the Video Game Description Language (VGDL) \cite{schaul2013video}. This is ideal for testing the generality of agents on multiple games or levels. GVG-AI includes over 100 classic arcade games each with five different levels.


\subsection{Racing Games}

Racing games are games where the player is tasked with controlling some kind of vehicle or character so as to reach a goal in the shortest possible time, or as to traverse as far as possible along a track in a given time. Usually, the game employs a first-person perspective or a vantage point from just behind the player-controlled vehicle. The vast majority of racing games take a continuous input signal as steering input, similar to a steering wheel. Some games, such as those in the \emph{Forza Motorsport} (Microsoft Studios, 2005--2016) or \emph{Real Racing} (Firemint and EA Games, 2009--2013) series, allow for complex input including gear stick, clutch and handbrake, whereas more arcade-focused games such as those in the \emph{Need for Speed} (Electronic Arts, 1994--2015) series typically have a simpler set of inputs and thus lower branching factor.

A challenge that is common in all racing games is that the agent needs to control the position of the vehicle and adjust the acceleration or braking, using fine-tuned continuous input, so as to traverse the track as fast as possible. Doing this optimally requires at least short-term planning, one or two turns forward. If there are resources to be managed in the game, such as fuel, damage or speed boosts, this requires longer-term planning. When other vehicles are present on the track, there is an adversarial planning aspect added, in trying to manage or block overtaking; this planning is often done in the presence of hidden information (position and resources of other vehicles on different parts of the track).

A popular environment for visual reinforcement learning with realistic 3D graphics is the open racing car simulator TORCS \cite{wymann2000torcs}.

\subsection{First-Person Shooters (FPS)}

More advanced game environments have recently emerged for visual reinforcement learning agents in a First-Person Shooters (FPS). In contrast to classic arcade games such as those in the ALE benchmark, FPSes have 3D graphics with partially observable states and are thus a more realistic environment to study. Usually, the viewpoint is that of the player-controlled character, though some games that are broadly in the FPS categories adopt an over-the-shoulder viewpoint. The design of FPS games is such that part of the challenge is simply fast perception and reaction, in particular, spotting enemies and quickly aiming at them. But there are other cognitive challenges as well, including orientation and movement in a complex three-dimensional environment, predicting actions and locations of multiple adversaries, and in some game modes also team-based collaboration. If visual inputs are used, there is the challenge of extracting relevant information from pixels.

Among FPS platforms are \emph{ViZDoom}, a framework that allows agents to play the classic first-person shooter Doom (id Software, 1993--2017) using the screen buffer as input \cite{kempka2016vizdoom}. \emph{DeepMind Lab} is a platform for 3D navigation and puzzle-solving tasks based on the \emph{Quake III Arena} (id Software, 1999) engine \cite{beattie2016deepmind}.

\subsection{Open-World Games}
Open-world games such as \emph{Minecraft} (Mojang, 2011) or the \emph{Grand Theft Auto} (Rockstar Games, 1997--2013) series are characterized by very non-linear gameplay, with a large game world to explore, either no set goals or many goals with unclear internal ordering, and large freedom of action at any given time. Key challenges for agents are exploring the world and setting goals which are realistic and meaningful. As this is a very complex challenge, most research use these open environments to explore reinforcement learning methods that can reuse and transfer learned knowledge to new tasks. Project Malmo is a platform built on top of the open-world game Minecraft, which can be used to define many diverse and complex problems \cite{johnson2016malmo}.

\subsection{Real-time Strategy Games}

Strategy games are games where the player controls multiple characters or units, and the objective of the game is to prevail in some sort of conquest or conflict. Usually, but not always, the narrative and graphics reflect a military conflict, where units may be e.g.\ knights, tanks or battleships. The key challenge in strategy games is to lay out and execute complex plans involving multiple units. This challenge is in general significantly harder than the planning challenge in classic board games such as Chess mainly because multiple units must be moved at any time and the effective branching factor is typically enormous. The planning horizon can be extremely long, where actions that are taken at the beginning of a game impact the overall strategy. In addition, there is the challenge of predicting the moves of one or several adversaries, who have multiple units themselves. Real-time Strategy Games (RTS) are strategy games which do not progress in discrete turns, but where actions can be taken at any point in time. RTS games add the challenge of time prioritization to the already substantial challenges of playing strategy games.

The \emph{StarCraft} (Blizzard Entertainment, 1998--2017) series is without a doubt the most studied game in the Real-Time Strategy (RTS) genre. The Brood War API (BWAPI)\footnote{http://bwapi.github.io/} enables software to communicate with StarCraft while the game runs, e.g.\ to extract state features and perform actions. BWAPI has been used extensively in game AI research, but currently, only a few examples exist where deep learning has been applied. TorchCraft is a library built on top of BWAPI that connects the scientific computing framework Torch to StarCraft to enable machine learning research for this game~\cite{synnaeve2016torchcraft}. Additionally, DeepMind and Blizzard (the developers of StarCraft) have developed a machine learning API to support research in StarCraft II with features such as simplified visuals designed for convolutional networks \cite{vinyals2017starcraft}. This API contains several mini-challenges while it also supports the full 1v1 game setting. \emph{$\mu$RTS} \cite{ontanon2013combinatorial} and \emph{ELF} \cite{tian2017elf} are two minimalistic RTS game engines that implement some of the features that are present in RTS games.

\subsection{Team Sports Games}
Popular sports games are typically based on team-based sports such as soccer, basketball, and football. These games aim to be as realistic as possible with life-like animations and 3D graphics.   Several soccer-like environments have been used extensively as research platforms, both with physical robots and 2D/3D simulations, in the annual Robot World Cup Soccer Games (RoboCup) \cite{asada2000overview}. 
\emph{Keepaway Soccer} is a simplistic soccer-like environment where one team of agents try to maintain control of the ball while another team tries to gain control of it \cite{stone2001keepaway}. A similar environment for multi-agent learning is \emph{RoboCup 2D Half-Field-Offense (HFO)} where teams of 2-3 players either take the role as offense or defense on one half of a soccer field \cite{hausknecht2016half}.

\subsection{Text Adventure Games}
A classic text adventure game is a form of interactive fiction where players are given descriptions and instructions in text, rather than graphics, and interact with the storyline through text-based commands \cite{sweetser2008emergence}. These commands are usually used to query the system about the state, interact with characters in the story, collect and use items, or navigate the space in the fictional world.

These games typically implement one of three text-based interfaces: parser-based, choice-Based, and hyperlink-based \cite{he2016drrn}. Choice-based and hyperlink-based interfaces provide the possible actions to the player at a given state as a list, out of context, or as links in the state description. Parser-Based interfaces are, on the other hand, open to any input and the player has to learn what words the game understands. This is interesting for computers as it is much more akin to natural language, where you have to know what actions should exist based on your understanding of language and the given state.

Unlike some other game genres, like arcade games, text adventure games have not had a standard benchmark of games that everyone can compare against. 
This makes a lot of results hard to directly compare. A lot of research has focused on games that run on Infocom's Z-Machine game engine, an engine that can play a lot of the early, classic games. Recently, Microsoft has introduced the environment TextWorld to help create a standardized text adventure environment \cite{cote18textworld}.

\subsection{OpenAI Gym \& Universe}
\emph{OpenAI Gym} is a large platform for comparing reinforcement learning algorithms with a single interface to a suite of different environments including ALE, GVG-AI, MuJoCo, Malmo, ViZDoom and more \cite{brockman2016openai}. OpenAI Universe is an extension to OpenAI Gym that currently interfaces with more than a thousand Flash games and aims to add many modern video games in the future\footnote{https://universe.openai.com/}.

\section{Deep Learning Methods for Game Playing}
\label{sec:game_playing}

\begin{figure*}[t!]
\begin{center}
  \includegraphics[width=\textwidth]{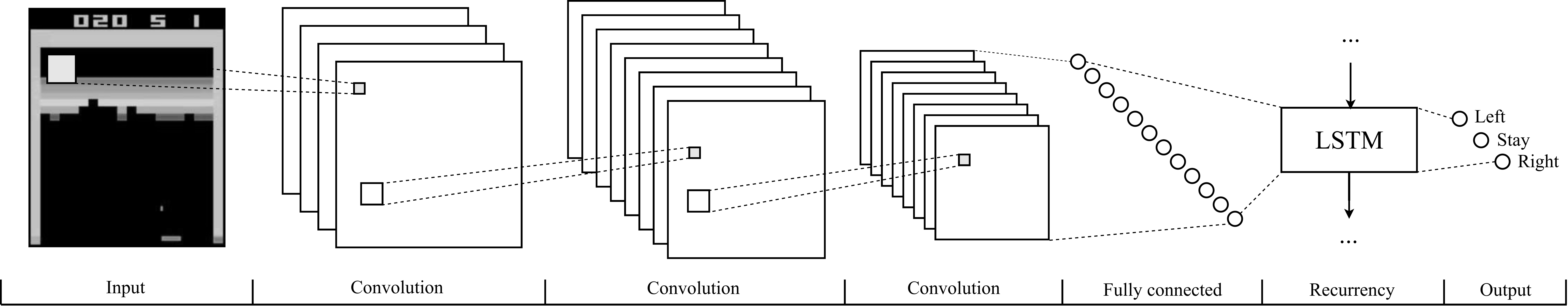} 
  \caption{An example of a typical network architecture used in deep reinforcement learning for game-playing with pixel input. The input usually consists of a preprocessed screen image, or several stacked or concatenated images, which is followed by a couple of convolutional layers (often without pooling), and a few fully connected layers. Recurrent networks have a recurrent layer, such as LSTM or GRU, after the fully connected layers. The output typically consists of one unit for each unique combination of actions in the game, and actor-critic methods also have one for the state value $V(s)$. Examples of this architecture, without a recurrent layer and with some variations, are \cite{mnih2013playing, mnih2015human, nair2015massively, van2016deep, schaul2015prioritized, osband2016deep, mnih2016asynchronous, wang2016sample, rusu2016progressive, salimans2017evolution, bellemare2017distributional, fortunato2018noisy, wang2015dueling, hessel2018rainbow, wu2017scalable, such2017deep, conti2018improving, espeholt2018impala}, and examples with a recurrent layer are \cite{hausknecht2015deep, mnih2016asynchronous, jaderberg2016reinforcement}. } 
  \label{fig:network}
  \vspace{-1.0em}
\end{center}
\end{figure*}

This section gives an overview of deep learning techniques used to play video games, divided by game genre. Table~\ref{tab:overview} lists deep learning methods for each game genre and highlights which input features, network architecture, and training methods they rely upon. A typical neural network architecture used in deep RL is shown in Figure~\ref{fig:network}. 


\subsection{Arcade Games}

The Arcade Learning Environment (ALE) consists of more than 50 Atari games and has been the main testbed for deep reinforcement learning algorithms that learn control policies directly from raw pixels. This section reviews the main advancements that have been demonstrated in ALE. An overview of these advancements is shown in Table \ref{tab:atari}. 

Deep Q-Network (DQN) was the first learning algorithm that showed human expert-level control in ALE \cite{mnih2013playing}. DQN was tested in seven Atari 2600 games and outperformed previous approaches, such as SARSA with feature construction \cite{bellemare2015arcade} and neuroevolution \cite{hausknecht2014neuroevolution}, as well as a human expert on three of the games. DQN is based on Q-learning, where a neural network model learns to approximate $Q^{\pi}(s,a)$ that estimates the expected return of taking action $a$ in state $s$ while following a behavior policy $\mu$. A simple network architecture consisting of two convolutional layers followed by a single fully-connected layer was used as a function approximator. 

A key mechanism in DQN is \emph{experience replay} \cite{lin1993reinforcement}, where experiences in the form $\{s_{t}, a_{t}, r_{t+1}, s_{t+1}\}$ are stored in a replay memory and randomly sampled in batches when the network is updated. This enables the algorithm to reuse and learn from past and uncorrelated experiences, which reduces the variance of the updates. DQN was later extended with a separate target Q-network which parameters are held fixed between individual updates and was shown to achieve above human expert scores in 29 out of 49 tested games \cite{mnih2015human}. 

Deep Recurrent Q-Learning (DRQN) extends the DQN architecture with a recurrent layer before the output and works well for games with partially observable states \cite{hausknecht2015deep}.

A distributed version of DQN was shown to outperform a non-distributed version in 41 of the 49 games using the Gorila architecture (General Reinforcement Learning Architecture) \cite{nair2015massively}. Gorila parallelizes actors that collect experiences into a distributed replay memory as well as parallelizing learners that train on samples from the same replay memory. 

One problem with the Q-learning algorithm is that it often overestimates action values because it uses the same value function for action-selection and action-evaluation. Double DQN, based on double Q-learning \cite{hasselt2010double}, reduces the observed overestimation by learning two value networks with parameters $\theta$ and $\theta^{'}$ that both use the other network for value-estimation, such that the target $Y_{t} = R_{t+1} + \gamma Q(S_{t+1}, \max_{a} Q(S_{t+1},a;\theta_{t});\theta^{'}_{t})$ \cite{van2016deep}. 

Another improvement is \emph{prioritized experience replay} from which important experiences are sampled more frequently based on the TD-error, which was shown to significantly improve both DQN and Double DQN \cite{schaul2015prioritized}.

Dueling DQN uses a network that is split into two streams after the convolutional layers to separately estimate state-value $V^{\pi}(s)$ and the action-advantage $A^{\pi}(s,a)$, such that $Q^{\pi}(s,a) = V^{\pi}(s) + A^{\pi}(s,a)$ \cite{wang2015dueling}. Dueling DQN improves Double DQN and can also be combined with prioritized experience replay. 

Double DQN and Dueling DQN were also tested in the five more complex games in the RLE and achieved a mean score of around 50\% of a human expert \cite{bhonker2016playing}. The best result in these experiments was by Dueling DQN in the game Mortal Kombat (Midway, 1992) with 128\%. 

Bootstrapped DQN improves exploration by training multiple Q-networks. A randomly sampled network is used during each training episode and \emph{bootstrap masks} modulate the gradients to train the networks differently \cite{osband2016deep}. 

Robust policies can be learned with DQN for competitive or cooperative multi-player games by training one network for each player and play them against each other in the training process \cite{tampuu2017multiagent}. Agents trained in multiplayer mode perform very well against novel opponents, whereas agents trained against a stationary algorithm fail to generalize their strategies to novel adversaries.

Multi-threaded asynchronous variants of DQN, SARSA and Actor-Critic methods can utilize multiple CPU threads on a single machine, reducing  training roughly linear to the number of parallel threads \cite{mnih2016asynchronous}. These variants do not rely on a replay memory because the network is updated on uncorrelated experiences from parallel actors which also helps to stabilize on-policy methods. The Asynchronous Advantage Actor-Critic (A3C) algorithm is an actor-critic method that uses several parallel agents to collect experiences that all asynchronously update a global actor-critic network. A3C outperformed Prioritized Dueling DQN, which was trained for 8 days on a GPU, with just half the training time on a CPU \cite{mnih2016asynchronous}.

An actor-critic method with experience replay (ACER) implements an efficient trust region policy method that forces updates to not deviate far from a running average of past policies \cite{wang2016sample}. The performance of ACER in ALE matches Dueling DQN with prioritized experience replay and A3C without experience replay, while it is much more data efficient. 

A3C with progressive neural networks \cite{rusu2016progressive} can effectively transfer learning from one game to another. The training is done by instantiating a network for every new task with connections to all the previous learned networks. This gives the new network access to knowledge already learned. 

The \emph{Advantage Actor-Critic} (A2C), a synchronous variant of A3C \cite{mnih2016asynchronous}, updates the parameters synchronously in batches and has comparable performance while only maintaining one neural network \cite{wu2017scalable}. Actor-Critic using Kronecker-Factored Trust Region (ACKTR) extends A2C by approximating the natural policy gradient updates for both the actor and the critic \cite{wu2017scalable}. In Atari, ACKTR has slower updates compared to A2C (at most 25\% per time step) but is more sample efficient (e.g.\ by a factor of 10 in Atlantis) \cite{wu2017scalable}. Trust Region Policy Optimization (TRPO) uses a \emph{surrogate} objective with theoretical guarantees for monotonic policy improvement, while it practically implements an approximation called \emph{trust region} \cite{schulman2015trust}. This is done by constraining network updates with a bound on the KL divergence between the current and the updated policy. TRPO has robust and data efficient performance in Atari games while it has high memory requirements and several restrictions. Proximal Policy Optimization (PPO) is an improvement on TRPO that uses a similar \emph{surrogate} objective \cite{schulman2017proximal}, but instead uses a soft constraint (originally suggested in \cite{schulman2015trust}) by adding the KL-divergence as a penalty. Instead of having a fixed penalty coefficient, it uses a clipped surrogate objective that penalizes policy updates outside some specified interval. PPO was shown to be more sample efficient than A2C and on par with ACER in Atari, while PPO does not rely on replay memory. PPO was also shown to have comparable or better performance than TRPO in continuous control tasks while being simpler and easier to parallelize.

\emph{IMPALA} (Importance Weighted Actor-Learner Architecture) is an actor-critic method where multiple learners with GPU access share gradients between each other while being synchronously updated from a set of actors \cite{espeholt2018impala}. This method can scale to a large number of machines and outperforms A3C. Additionally, IMPALA was trained, with one set of parameters, to play all 57 Atari games in ALE with a mean human-normalized score of  176.9\% (median of 59.7\%) \cite{espeholt2018impala}. Experiences collected by the actors in the IMPALA setup can lack behind the learners' policy and thus result in off-policy learning. This discrepancy is mitigated through a \emph{V-trace} algorithm that weighs the importance of experiences based on the difference between the actor's and learner's policies \cite{espeholt2018impala}.

\emph{UNREAL} (UNsupervised REinforcement and Auxiliary Learning) algorithm is based on A3C but uses a replay memory from which it learns auxiliary tasks and \emph{pseudo-reward functions} concurrently \cite{jaderberg2016reinforcement}. UNREAL only shows a small improvement over vanilla A3C in ALE, but larger improvements in other domains (see Section~\ref{sec:fps}). 

\emph{Distributional DQN} takes a distributional perspective on reinforcement learning by treating $Q(s,a)$ as an approximate distribution of returns instead of a single approximate expectation for each action \cite{bellemare2017distributional}. The distribution is divided into a so-called set of atoms, which determines the granularity of the distribution. Their results show that the more fine-grained the distributions are, the better are the results, and with 51 atoms (this variant was called C51) it achieved mean scores in ALE almost comparable to UNREAL.

In \emph{NoisyNets}, noise is added to the network parameters and a unique noise level for each parameter is learned using gradient descent \cite{fortunato2018noisy}. In contrast to $\epsilon$-greedy exploration, where an agent either samples actions from the policy or from a uniform random distribution, NoisyNets use a noisy version of the policy to ensure exploration, and this was shown to improve DQN (NoisyNet-DQN) and A3C (NoisyNet-A3C).

\emph{Rainbow} combines several DQN enhancements: Double DQN, Prioritized Replay, Dueling DQN, Distributional DQN, and NoisyNets, and achieved a mean score higher than any of the enhancements individually \cite{hessel2018rainbow}.

\emph{Evolution Strategies} (ES) are black-box optimization algorithms that rely on parameter-exploration through stochastic noise instead of calculating gradients and were found to be highly parallelizable with a linear speedup in training time when more CPUs are used \cite{salimans2017evolution}. 720 CPUs were used for one hour whereafter ES managed to outperform A3C (which ran for 4 days) in 23 out of 51 games, while ES used 3 to 10 times as much data due to its high parallelization. ES only ran a single day and thus their full potential is currently unknown. Novelty search is a popular algorithm that can overcome environments with deceptive and/or sparse rewards by guiding the search towards novel behaviors \cite{lehman2008exploiting}. ES has been extended to use Novelty Search (NS-ES) which outperforms ES on several challenging Atari games by defining novel behaviors based on the RAM states \cite{conti2018improving}. A quality-diversity variant called NSR-ES that uses both novelty and the reward signal reach an even higher performance \cite{conti2018improving}. NS-ES and NSR-ES reached worse results on a few games, possibly where the reward function is not sparse or deceptive. 

A simple genetic algorithm with a Gaussian noise mutation operator evolves the parameters of a deep neural network (Deep GA) and can achieve surprisingly good scores across several Atari games \cite{such2017deep}. Deep GA shows comparable results to DQN, A3C, and ES on 13 Atari games using up to thousands of CPUs in parallel. Additionally, random search, given roughly the same amount of computation, was shown to outperform DQN on 4 out of 13 games and A3C on 5 games \cite{such2017deep}. While there has been concern that evolutionary methods do not scale as well as gradient descent-based methods, one possibility is separating the feature construction from the policy network; evolutionary algorithms can then create extremely small networks that still play well~\cite{cuccu2018playing}.


\begin{table}[t]
  \centering
  \begin{tabular}{ l|r|r|c }
    Results & Mean & Median & Year and orig. paper\\ 
    \hline
    \hline
    DQN \cite{wang2015dueling} & 228\% & 79\% & 2013 \cite{mnih2013playing} 
    \\
    Double DQN (DDQN) \cite{wang2015dueling}& 307\% & 118\% & 2015 \cite{van2016deep} 
    \\
    Dueling DDQN \cite{wang2015dueling} & 373\% & 151\% & 2015 \cite{wang2015dueling} 
    \\
    Prior. DDQN \cite{wang2015dueling} & 435\% & 124\% & 2015 \cite{schaul2015prioritized} 
    \\
    Prior. Duel DDQN \cite{wang2015dueling} & 592\% & 172\% & 2015 \cite{schaul2015prioritized} 
    \\
    A3C \cite{jaderberg2016reinforcement} & 853\% & N/A & 2016 \cite{mnih2016asynchronous} 
    \\
    UNREAL \cite{jaderberg2016reinforcement}* & 880\% & 250\% & 2016 \cite{jaderberg2016reinforcement} 
    \\
    NoisyNet-DQN \cite{hessel2018rainbow} & N/A & 118\% & 2017 \cite{fortunato2018noisy} 
    \\
    Distr. DQN (C51) \cite{bellemare2017distributional} & 701\% & 178\% & 2017 \cite{bellemare2017distributional} 
    \\
    Rainbow \cite{hessel2018rainbow} & N/A & 223\% & 2017 \cite{hessel2018rainbow} 
    \\
    IMPALA \cite{espeholt2018impala} & 958\% & 192\% & 2018 \cite{espeholt2018impala} 
    \\
    Ape-X DQN \cite{horgan2018distributed} & N/A & 434\% & 2018 \cite{horgan2018distributed} 
    \\
    \hline
  \end{tabular}
  \label{tab:atari}
\caption{Human-normalized scores reported with various deep reinforcement learning algorithms in ALE on 57 Atari games using the \emph{30 no-ops} evaluation metric. References in the first column refer to the paper that included the results, while the last column references the paper that first introduced the specific technique. Note, that the reported scores use various amounts of training time and resources, thus not entirely comparable. Successors typically use more resources and less wall-clock time. *Hyper-parameters was tuned for every game leading to higher scores for UNREAL.}
\label{tab:deep-reinforcement-learning-methods}
\vspace{-2.0em}
\end{table}

A few supervised learning approaches have been applied to arcade games. In Guo et al.~\cite{guo2014deep} a slow planning agent was applied offline, using Monte-Carlo Tree Search, to generate data for training a CNN via multinomial classification. This approach, called \emph{UCTtoClassification}, was shown to outperform DQN. Policy distillation~\cite{rusu2015policy} or actor-mimic~\cite{parisotto2015actor} methods can be used to train one network to mimic a set of policies (e.g. for different games). These methods can reduce the size of the network and sometimes also improve the performance. A frame prediction model can be learned from a dataset generated by a DQN agent using the encoding-transformation-decoding network architecture; the model can then be used to improve exploration in a retraining phase~\cite{oh2015action}. Self-supervised tasks, such as reward prediction, validation of state-successor pairs, and mapping states and successor states to actions can define auxiliary losses used in pre-training of a policy network, which ultimately can improve learning \cite{shelhamer2016loss}.



The \emph{training objective} provides feedback to the agent while the \emph{performance objective} specifies the target behavior. Often, a single reward function takes both roles, but for some games, the performance objective does not guide the training sufficiently. The Hybrid Reward Architecture (HRA) splits the reward function into $n$ different reward functions, where each of them are assigned a separate learning agent \cite{van2017hybrid}. HRA does this by having $n$ output streams in the network, and thus $n$ Q-values, which are combined when actions are selected. HRA was able to achieve the maximum possible score in less than 3,000 episodes.

\subsection{Montezuma's Revenge}
\label{sec:montezuma}
Environments with sparse feedback remain an open challenge for reinforcement learning. The game \emph{Montezuma's Revenge} is a good example of such an environment in ALE and has thus been studied in more detail and used for benchmarking learning methods based on intrinsic motivation and curiosity. The main idea of applying intrinsic motivation is to improve the exploration of the environment based on some self-rewarding system, which eventually will help the agent to obtain an extrinsic reward. DQN fails to obtain any reward in this game (receiving a score of 0) and Gorila achieves an average score of just 4.2. A human expert can achieve 4,367 points and it is clear that the methods presented so far are unable to deal with environments with such sparse rewards. A few promising methods aim to overcome these challenges.

Hierarchical-DQN (h-DQN) \cite{kulkarni2016hierarchical}  operates on two temporal scales, where one Q-value function $Q_{1}(s,a;g)$, the \emph{controller}, learns a policy over actions that satisfy goals chosen by a higher-level Q-value function $Q_{2}(s, g)$, the \emph{meta-controller}, which learns a policy over intrinsic goals (i.e.\ which goals to select). This method was able to reach an average score of around 400 in Montezuma's Revenge where goals were defined as states in which the agent \emph{reaches} (collides with) a certain type of object. This method, therefore, must rely on some object detection mechanism. 

Pseudo-counts have been used to provide intrinsic motivation in the form of exploration bonuses when unexpected pixel configurations are observed and can be derived from CTS density models \cite{bellemare2016unifying} or neural density models \cite{ostrovski2017count}. Density models assign probabilities to images, and a model's pseudo count of an observed image is the model's change in prediction compared to being trained one  additional time on the same image.
Impressive results were achieved in Montezuma's Revenge and other hard Atari games by combining DQN with the CTS density model (DQN-CTS) or the PixelCNN density model (DQN-PixelCNN) \cite{bellemare2016unifying}. Interestingly, the results were less impressive when the CTS density model was combined with A3C (A3C-CTS) \cite{bellemare2016unifying}. 

\begin{table*}[t!]
  \centering
  \caption{Overview of deep learning methods applied to games. We refer to \emph{features} as low-dimensional items and values that describe the state of the game such as health, ammunition, score, objects, etc. \emph{MLP} refers to a traditional fully-connected architecture without convolutional or recurrent layers. }
\setlength\tabcolsep{3.6pt}
  \begin{tabular}{ clcc*4lcc*4lcc*3lcc*4lcc*7lc }
    \Xhline{2\arrayrulewidth}
    
    Game & Method 
    & & & \multicolumn{3}{c}{Input} & & & \multicolumn{4}{c}{Architecture} 
    & & & \multicolumn{5}{c}{\specialcell{Training}} 
    & & & \multicolumn{5}{c}{Miscellaneous} 
    \\
    
    \hline
     
    \multirow{26}{*}{\specialcell{Atari\\2600\\(ALE)}} & 
    & & & \mcrot{1}{l}{60}{Features} & \mcrot{1}{l}{60}{Pixels} & \mcrot{1}{l}{60}{Text} 
    & & & \mcrot{1}{l}{60}{CNN} & \mcrot{1}{l}{60}{RNN} & \mcrot{1}{l}{60}{Ext. Memory} & \mcrot{1}{l}{60}{MLP} 
    & & & \mcrot{1}{l}{60}{Supervised} & \mcrot{1}{l}{60}{Q-learning} & \mcrot{1}{l}{60}{Actor-critic} & \mcrot{1}{l}{60}{ES} & \mcrot{1}{l}{60}{GA} 
    & & & \mcrot{1}{l}{60}{Auxiliary Learning} & \mcrot{1}{l}{60}{Hierarchical} & \mcrot{1}{l}{60}{Intrinsic Motivation} & \mcrot{1}{l}{60}{Transfer Learning} & \mcrot{1}{l}{60}{Distributed}
    & \\ 
    
    \hline
    
    & DQN \cite{mnih2013playing, mnih2015human} 
    & & & \fullmoon & \newmoon & \fullmoon 
    & & & \newmoon & \fullmoon & \fullmoon & \fullmoon 
    & & & \fullmoon & \newmoon & \fullmoon & \fullmoon & \fullmoon 
    & & & \fullmoon &\fullmoon & \fullmoon & \fullmoon & \fullmoon 
    & \\
    
    & DRQN \cite{hausknecht2015deep}
    & & & \fullmoon & \newmoon & \fullmoon 
    & & & \newmoon & \newmoon & \fullmoon & \fullmoon 
    & & & \fullmoon & \newmoon & \fullmoon & \fullmoon & \fullmoon 
    & & & \fullmoon & \fullmoon & \fullmoon & \fullmoon & \fullmoon 
    & \\
    
    & UCTtoClassification \cite{guo2014deep}
    & & & \fullmoon & \newmoon & \fullmoon 
    & & & \newmoon & \fullmoon & \fullmoon & \fullmoon 
    & & & \newmoon & \fullmoon & \fullmoon & \fullmoon & \fullmoon 
    & & & \fullmoon & \fullmoon & \fullmoon & \fullmoon & \fullmoon 
    & \\
    
    & Gorila \cite{nair2015massively}
    & & & \fullmoon & \newmoon & \fullmoon 
    & & & \newmoon & \fullmoon & \fullmoon & \fullmoon 
    & & & \fullmoon & \newmoon & \fullmoon & \fullmoon & \fullmoon 
    & & & \fullmoon & \fullmoon & \fullmoon & \fullmoon & \newmoon 
    & \\
    
    & Double DQN \cite{van2016deep}
    & & & \fullmoon & \newmoon & \fullmoon 
    & & & \newmoon & \fullmoon & \fullmoon & \fullmoon 
    & & & \fullmoon & \newmoon & \fullmoon & \fullmoon & \fullmoon 
    & & & \fullmoon & \fullmoon & \fullmoon & \fullmoon & \fullmoon 
    & \\
    
    & Prioritized DQN \cite{schaul2015prioritized}
    & & & \fullmoon & \newmoon & \fullmoon 
    & & & \newmoon & \fullmoon & \fullmoon & \fullmoon 
    & & & \fullmoon & \newmoon & \fullmoon & \fullmoon & \fullmoon 
    & & & \fullmoon & \fullmoon & \fullmoon & \fullmoon & \fullmoon 
    & \\
    
    & Dueling DQN \cite{wang2015dueling}
    & & & \fullmoon & \newmoon & \fullmoon 
    & & & \newmoon & \fullmoon & \fullmoon & \fullmoon 
    & & & \fullmoon & \newmoon & \fullmoon & \fullmoon & \fullmoon 
    & & & \fullmoon & \fullmoon & \fullmoon & \fullmoon & \fullmoon 
    & \\
    
    & Bootstrapped DQN \cite{osband2016deep}
    & & & \fullmoon & \newmoon & \fullmoon 
    & & & \newmoon & \fullmoon & \fullmoon & \fullmoon 
    & & & \fullmoon & \newmoon & \fullmoon & \fullmoon & \fullmoon 
    & & & \fullmoon & \fullmoon & \fullmoon & \fullmoon & \fullmoon 
    & \\
    
    & A3C / A2C \cite{mnih2016asynchronous}
    & & & \fullmoon & \newmoon & \fullmoon 
    & & & \newmoon & \newmoon & \fullmoon & \fullmoon 
    & & & \fullmoon & \fullmoon & \newmoon & \fullmoon & \fullmoon 
    & & & \fullmoon & \fullmoon & \fullmoon & \fullmoon & \newmoon 
    & \\
    
  	& ACER \cite{wang2016sample}
    & & & \fullmoon & \newmoon & \fullmoon 
    & & & \newmoon & \fullmoon & \fullmoon & \fullmoon 
    & & & \fullmoon & \fullmoon & \newmoon & \fullmoon & \fullmoon 
    & & & \fullmoon & \fullmoon & \fullmoon & \fullmoon & \newmoon 
    & \\
    
    & Progressive Networks \cite{rusu2016progressive}
    & & & \fullmoon & \newmoon & \fullmoon 
    & & & \newmoon & \fullmoon & \fullmoon & \fullmoon 
    & & & \fullmoon & \fullmoon & \newmoon & \fullmoon & \fullmoon 
    & & & \fullmoon & \fullmoon & \fullmoon & \newmoon & \newmoon
    & \\
    
    & UNREAL \cite{jaderberg2016reinforcement} 
    & & & \fullmoon & \newmoon & \fullmoon 
    & & & \newmoon & \newmoon & \fullmoon & \fullmoon 
    & & & \fullmoon & \fullmoon & \newmoon & \fullmoon & \fullmoon 
    & & & \newmoon & \fullmoon & \fullmoon & \fullmoon & \newmoon 
    & \\
    
    & Scalable Evolution Strategies \cite{salimans2017evolution}
    & & & \fullmoon & \newmoon & \fullmoon 
    & & & \newmoon & \fullmoon & \fullmoon & \fullmoon 
    & & & \fullmoon & \fullmoon & \fullmoon & \newmoon & \fullmoon 
    & & & \fullmoon & \fullmoon & \fullmoon & \fullmoon & \newmoon 
    & \\
    
    & Distributional DQN (C51) \cite{bellemare2017distributional}
    & & & \fullmoon & \newmoon & \fullmoon 
    & & & \newmoon & \fullmoon & \fullmoon & \fullmoon 
    & & & \fullmoon & \newmoon & \fullmoon & \fullmoon & \fullmoon 
    & & & \fullmoon & \fullmoon & \fullmoon & \fullmoon & \fullmoon 
    & \\
    
    & NoisyNet-DQN \cite{fortunato2018noisy}
    & & & \fullmoon & \newmoon & \fullmoon 
    & & & \newmoon & \fullmoon & \fullmoon & \fullmoon 
    & & & \fullmoon & \newmoon & \fullmoon & \fullmoon & \fullmoon 
    & & & \fullmoon & \fullmoon & \fullmoon & \fullmoon & \fullmoon
    & \\
    
    & NoisyNet-A3C \cite{fortunato2018noisy}
    & & & \fullmoon & \newmoon & \fullmoon 
    & & & \newmoon & \fullmoon & \fullmoon & \fullmoon 
    & & & \fullmoon & \fullmoon & \newmoon & \fullmoon & \fullmoon 
    & & & \fullmoon & \fullmoon & \fullmoon & \fullmoon & \newmoon
    & \\
    
    & Rainbow \cite{hessel2018rainbow}
    & & & \fullmoon & \newmoon & \fullmoon 
    & & & \newmoon & \fullmoon & \fullmoon & \fullmoon 
    & & & \fullmoon & \newmoon & \fullmoon & \fullmoon & \fullmoon 
    & & & \fullmoon & \fullmoon & \fullmoon & \fullmoon & \fullmoon 
    & \\
    
    & ACKTR \cite{wu2017scalable}
    & & & \fullmoon & \newmoon & \fullmoon 
    & & & \newmoon & \fullmoon & \fullmoon & \fullmoon 
    & & & \fullmoon & \fullmoon & \newmoon & \fullmoon & \fullmoon 
    & & & \fullmoon & \fullmoon & \fullmoon & \fullmoon & \newmoon 
    & \\
    
    & Deep GA \cite{such2017deep}
    & & & \fullmoon & \newmoon & \fullmoon 
    & & & \newmoon & \fullmoon & \fullmoon & \fullmoon 
    & & & \fullmoon & \fullmoon & \fullmoon & \fullmoon & \newmoon 
    & & & \fullmoon & \fullmoon & \fullmoon & \fullmoon & \newmoon 
    & \\
    
    & NS-ES / NSR-ES \cite{conti2018improving}
    & & & \fullmoon & \newmoon & \fullmoon 
    & & & \newmoon & \fullmoon & \fullmoon & \fullmoon 
    & & & \fullmoon & \fullmoon & \fullmoon & \newmoon & \fullmoon 
    & & & \fullmoon & \fullmoon & \fullmoon & \fullmoon & \newmoon 
    & \\
    
    & IMPALA \cite{espeholt2018impala}
    & & & \fullmoon & \newmoon & \fullmoon 
    & & & \newmoon & \fullmoon & \fullmoon & \fullmoon 
    & & & \fullmoon & \fullmoon & \newmoon & \fullmoon & \fullmoon 
    & & & \fullmoon & \fullmoon & \fullmoon & \fullmoon & \newmoon 
    & \\
    
    & TRPO \cite{schulman2015trust}
    & & & \fullmoon & \newmoon & \fullmoon 
    & & & \newmoon & \fullmoon & \fullmoon & \fullmoon 
    & & & \fullmoon & \fullmoon & \newmoon & \fullmoon & \fullmoon 
    & & & \fullmoon & \fullmoon & \fullmoon & \fullmoon & \newmoon 
    & \\
    
    & PPO \cite{schulman2017proximal}
    & & & \fullmoon & \newmoon & \fullmoon 
    & & & \newmoon & \newmoon & \fullmoon & \fullmoon 
    & & & \fullmoon & \fullmoon & \newmoon & \fullmoon & \fullmoon 
    & & & \fullmoon & \fullmoon & \fullmoon & \fullmoon & \newmoon 
    & \\
    
    & 
    DQfD \cite{hester2017deep}
    & & & \fullmoon & \newmoon & \fullmoon 
    & & & \newmoon & \fullmoon & \fullmoon & \fullmoon 
    & & & \newmoon & \newmoon & \fullmoon & \fullmoon & \fullmoon 
    & & & \fullmoon & \fullmoon & \fullmoon & \fullmoon & \fullmoon 
    & \\
     
    & Ape-X DQN \cite{horgan2018distributed}
    & & & \fullmoon & \newmoon & \fullmoon 
    & & & \newmoon & \fullmoon & \fullmoon & \fullmoon 
    & & & \fullmoon & \newmoon & \fullmoon & \fullmoon & \fullmoon 
    & & & \fullmoon & \fullmoon & \fullmoon & \fullmoon & \newmoon 
    & \\
    
    & 
    Ape-X DQfD \cite{pohlen2018observe}
    & & & \fullmoon & \newmoon & \fullmoon 
    & & & \newmoon & \fullmoon & \fullmoon & \fullmoon 
    & & & \newmoon & \newmoon & \fullmoon & \fullmoon & \fullmoon 
    & & & \fullmoon & \fullmoon & \fullmoon & \fullmoon & \newmoon 
    & \\
    
    \cline{2-30}
    
	\multirow{1}{*}{\specialcell{Ms. Pac-Man}} & 
    Hybrid Reward Architecture (HRA) \cite{van2017hybrid}
    & & & \newmoon & \newmoon & \fullmoon 
    & & & \newmoon & \fullmoon & \fullmoon & \fullmoon 
    & & & \fullmoon & \newmoon & \fullmoon & \fullmoon & \fullmoon 
    & & & \fullmoon & \fullmoon & \fullmoon & \fullmoon & \fullmoon
    & \\
    
    \cline{2-30}
    
    \multirow{2}{*}{\specialcell{Montezuma's\\Revenge}} & 
    Hierarchical-DQN (h-DQN) \cite{kulkarni2016hierarchical}
    & & & \newmoon & \newmoon & \fullmoon 
    & & & \newmoon & \fullmoon & \fullmoon & \fullmoon 
    & & & \fullmoon & \newmoon & \fullmoon & \fullmoon & \fullmoon 
    & & & \fullmoon & \newmoon & \newmoon & \fullmoon & \fullmoon 
    & \\
    
    & 
    DQN-CTS / DQN-PixelCNN \cite{bellemare2016unifying}
    & & & \fullmoon & \newmoon & \fullmoon 
    & & & \newmoon & \fullmoon & \fullmoon & \fullmoon 
    & & & \fullmoon & \newmoon & \fullmoon & \fullmoon & \fullmoon 
    & & & \fullmoon & \fullmoon & \newmoon & \fullmoon & \fullmoon 
    & \\
    
    \hline
    
    \multirow{3}{*}{Racing} & 
    Direct Perception \cite{chen2015deepdriving}
    & & & \newmoon & \newmoon & \fullmoon 
    & & & \newmoon & \fullmoon & \fullmoon & \fullmoon 
    & & & \newmoon & \fullmoon & \fullmoon & \fullmoon & \fullmoon 
    & & & \fullmoon & \fullmoon & \fullmoon & \newmoon & \fullmoon 
    & \\
    
    & 
    Deep DPG (DDPG) \cite{lillicrap2015continuous}
    & & & \fullmoon & \newmoon & \fullmoon 
    & & & \newmoon & \fullmoon & \fullmoon & \fullmoon 
    & & & \fullmoon & \fullmoon & \newmoon & \fullmoon & \fullmoon 
    & & & \fullmoon & \fullmoon & \fullmoon & \fullmoon & \fullmoon 
    & \\
    
    & 
    A3C \cite{mnih2016asynchronous}
    & & & \fullmoon & \newmoon & \fullmoon 
    & & & \newmoon & \fullmoon & \fullmoon & \fullmoon 
    & & & \fullmoon & \fullmoon & \newmoon & \fullmoon & \fullmoon 
    & & & \fullmoon & \fullmoon & \fullmoon & \fullmoon & \newmoon 
    & \\
    
    \hline
    
    \multirow{5}{*}{Doom} & 
   	DQN \cite{kempka2016vizdoom}
    & & & \fullmoon & \newmoon & \fullmoon 
    & & & \newmoon & \fullmoon & \fullmoon & \fullmoon 
    & & & \fullmoon & \newmoon & \fullmoon & \fullmoon & \fullmoon
    & & & \fullmoon & \fullmoon & \fullmoon & \fullmoon & \fullmoon 
    & \\
    
    & 
   	A3C \cite{wu2017training}
    & & & \fullmoon & \newmoon & \fullmoon 
    & & & \newmoon & \fullmoon & \fullmoon & \fullmoon 
    & & & \fullmoon & \fullmoon & \newmoon & \fullmoon & \fullmoon 
    & & & \fullmoon & \fullmoon & \fullmoon & \fullmoon & \newmoon 
    & \\
    
    & 
   	DRQN \cite{lample2016playing}
    & & & \fullmoon & \newmoon & \fullmoon 
    & & & \newmoon & \newmoon & \fullmoon & \fullmoon 
    & & & \fullmoon & \newmoon & \fullmoon & \fullmoon & \fullmoon 
    & & & \newmoon & \fullmoon & \fullmoon & \fullmoon & \fullmoon 
    & \\
    
    & 
   	DQN + SLAM \cite{bhatti2016playing}
    & & & \fullmoon & \newmoon & \fullmoon 
    & & & \newmoon & \fullmoon & \fullmoon & \fullmoon 
    & & & \newmoon & \newmoon & \fullmoon & \fullmoon & \fullmoon 
    & & & \fullmoon & \fullmoon & \fullmoon & \fullmoon & \fullmoon 
    & \\
    
    & 
   	Direct Future Prediction (DFP) \cite{dosovitskiy2016learning}
    & & & \newmoon & \newmoon & \fullmoon 
    & & & \newmoon & \fullmoon & \fullmoon & \fullmoon 
    & & & \fullmoon & \fullmoon & \fullmoon & \fullmoon & \fullmoon 
    & & & \fullmoon & \fullmoon & \fullmoon & \newmoon & \newmoon 
    & \\
    
    \hline
    
    \multirow{3}{*}{Minecraft} & H-DRLN \cite{tessler2016deep}
    & & & \newmoon & \newmoon & \fullmoon 
    & & & \newmoon & \fullmoon & \fullmoon & \fullmoon 
    & & & \fullmoon & \newmoon & \fullmoon & \fullmoon & \fullmoon 
    & & & \fullmoon & \newmoon & \fullmoon & \fullmoon & \fullmoon 
    & \\
    
    & RMQN / FRMQN \cite{oh2016control}
    & & & \fullmoon & \newmoon & \fullmoon 
    & & & \newmoon & \newmoon & \newmoon & \fullmoon 
    & & & \fullmoon & \newmoon & \fullmoon & \fullmoon & \fullmoon 
    & & & \fullmoon & \fullmoon & \fullmoon & \newmoon & \fullmoon 
    & \\
    
    & TSCL \cite{matiisen2017teacher}
    & & & \fullmoon & \newmoon & \fullmoon 
    & & & \newmoon & \newmoon & \fullmoon & \fullmoon 
    & & & \fullmoon & \fullmoon & \fullmoon & \fullmoon & \fullmoon
    & & & \fullmoon & \fullmoon & \fullmoon & \fullmoon & \fullmoon 
    & \\
    
    \hline
    
    \multirow{6}{*}{StarCraft} & Zero Order \cite{usunier2016episodic}
    & & & \newmoon & \fullmoon & \fullmoon 
    & & & \fullmoon & \fullmoon & \fullmoon & \newmoon 
    & & & \fullmoon & \fullmoon & \fullmoon & \fullmoon & \fullmoon 
    & & & \fullmoon & \fullmoon & \fullmoon & \fullmoon & \fullmoon 
    & \\
    
    & IQL \cite{foerster2017stabilising}
    & & & \newmoon & \fullmoon & \fullmoon 
    & & & \fullmoon & \newmoon & \fullmoon & \fullmoon 
    & & & \fullmoon & \newmoon & \fullmoon & \fullmoon & \fullmoon 
    & & & \fullmoon & \fullmoon & \fullmoon & \fullmoon & \fullmoon 
    & \\
    
    & BiCNet \cite{peng2017multiagent}
    & & & \newmoon & \fullmoon & \fullmoon 
    & & & \fullmoon & \newmoon & \fullmoon & \fullmoon 
    & & & \fullmoon & \fullmoon & \newmoon & \fullmoon & \fullmoon 
    & & & \fullmoon & \fullmoon & \fullmoon & \fullmoon & \fullmoon 
    & \\
    
    & COMA \cite{foerster2017counterfactual}
    & & & \newmoon & \fullmoon & \fullmoon 
    & & & \fullmoon & \newmoon & \fullmoon & \fullmoon 
    & & & \fullmoon & \fullmoon & \newmoon & \fullmoon & \fullmoon 
    & & & \fullmoon & \fullmoon & \fullmoon & \fullmoon & \fullmoon 
    & \\
    
    & Macro-action SL \cite{justesen2017learning}
    & & & \newmoon & \fullmoon & \fullmoon 
    & & & \fullmoon & \fullmoon & \fullmoon & \newmoon 
    & & & \newmoon & \fullmoon & \fullmoon & \fullmoon & \fullmoon 
    & & & \fullmoon & \newmoon & \fullmoon & \fullmoon & \fullmoon
    & \\
    
    & Macro-action CNNFQ/PPO \cite{tang2018reinforcement}, \cite{sun2018tstarbots}
    & & & \newmoon & \newmoon & \fullmoon 
    & & & \newmoon & \fullmoon & \fullmoon & \newmoon 
    & & & \fullmoon & \fullmoon & \newmoon & \fullmoon & \fullmoon 
    & & & \fullmoon & \newmoon & \fullmoon & \fullmoon & \newmoon 
    & \\
    
    
    \hline
      
      \multirow{2}{*}{\specialcell{RoboCup Soccer\\(HFO)}} & DDPG + Inverting Gradients \cite{hausknecht2015deepparam}
    & & & \newmoon & \fullmoon & \fullmoon 
    & & & \fullmoon & \fullmoon & \fullmoon & \newmoon 
    & & & \fullmoon & \fullmoon & \newmoon & \fullmoon & \fullmoon 
    & & & \fullmoon & \fullmoon & \fullmoon & \fullmoon & \fullmoon 
    & \\
    
    & DDPG + Mixing policy targets \cite{hausknecht2016policy}
    & & & \newmoon & \fullmoon & \fullmoon 
    & & & \fullmoon & \fullmoon & \fullmoon & \newmoon 
    & & & \fullmoon & \fullmoon & \newmoon & \fullmoon & \fullmoon 
    & & & \fullmoon & \fullmoon & \fullmoon & \fullmoon & \fullmoon 
    & \\
    
    \hline
    
    \multirow{1}{*}{2D billiard} & Object-centric prediction \cite{fragkiadaki2015learning}
    & & & \newmoon & \newmoon & \fullmoon 
    & & & \newmoon & \newmoon & \fullmoon & \fullmoon 
    & & & \newmoon & \fullmoon & \fullmoon & \fullmoon & \fullmoon 
    & & & \fullmoon & \fullmoon & \fullmoon & \fullmoon & \fullmoon 
    & \\
    
    \hline
    
    \multirow{5}{*}{\specialcell{Text adventure\\games}} & LSTM-DQN \cite{narasimhan2015language}
    & & & \fullmoon & \fullmoon & \newmoon 
    & & & \fullmoon & \newmoon & \fullmoon & \fullmoon 
    & & & \fullmoon & \newmoon & \fullmoon & \fullmoon & \fullmoon 
    & & & \fullmoon & \fullmoon & \fullmoon & \fullmoon & \fullmoon 
    & \\
    
    & DRRN \cite{he2016drrn}
    & & & \fullmoon & \fullmoon & \newmoon 
    & & & \fullmoon & \fullmoon & \fullmoon & \newmoon 
    & & & \fullmoon & \newmoon & \fullmoon & \fullmoon & \fullmoon 
    & & & \newmoon & \fullmoon & \fullmoon & \fullmoon & \fullmoon 
    & \\
    
    & Affordance Based Action Selection \cite{fulda2017can}
    & & & \fullmoon & \fullmoon & \newmoon 
    & & & \fullmoon & \fullmoon & \fullmoon & \newmoon 
    & & & \fullmoon & \newmoon & \fullmoon & \fullmoon & \fullmoon 
    & & & \newmoon & \fullmoon & \fullmoon & \fullmoon & \fullmoon 
    & \\
    
    & Golovin \cite{kostka2017text}
    & & & \fullmoon & \fullmoon & \newmoon 
    & & & \fullmoon & \newmoon & \fullmoon & \fullmoon 
    & & & \fullmoon & \fullmoon & \fullmoon & \fullmoon & \fullmoon 
    & & & \fullmoon & \fullmoon & \fullmoon & \fullmoon & \fullmoon 
    & \\
    
    & AE-DQN \cite{zahavy2018learn}
    & & & \fullmoon & \fullmoon & \newmoon 
    & & & \newmoon & \fullmoon & \fullmoon & \fullmoon 
    & & & \fullmoon & \newmoon & \fullmoon & \fullmoon & \fullmoon 
    & & & \newmoon & \fullmoon & \fullmoon & \fullmoon & \fullmoon 
    & \\
      
  \Xhline{2\arrayrulewidth}
  \end{tabular}
\label{tab:overview}
\vspace{-1.5em}

\end{table*}

Ape-X DQN is a distributed DQN architecture similar to Gorila, as in actors are separated from the learner. Ape-X DQN was able to reach state-of-art results across the 57 Atari games using 376 cores and 1 GPU, running at ∼50K FPS~\cite{horgan2018distributed}. Deep Q-learning from Demonstrations (DQfD) draw samples from an experience replay buffer that is initialized with demonstration data from a human expert and is superior to previous methods on 11 Atari games with sparse rewards~\cite{hester2017deep}. Ape-X DQfD combines the distributed architecture from Ape-X and the learning algorithm from DQfD using expert data and was shown to outperform all previous methods in ALE as well as beating level 1 in Montezuma's Revenge~\cite{pohlen2018observe}.

To improve the performance, Kaplan et. al. augmented the agent training with text instructions. An instruction-based reinforcement learning approach that uses both a CNN for visual input and RNN for text-based instruction, inputs managed to achieve a score of 3,500 points. Instructions were linked to positions in rooms and agents were rewarded when they reached those locations \cite{kaplan2017beating}, demonstrating a fruitful collaboration between a human and a learning algorithm. Experiments in Montezuma's Revenge also showed that the network learned to generalize to unseen instructions that were similar to previous instructions.

Similar work demonstrates how an agent can execute text-based commands in a 2D maze-like environment called XWORLD, such as walking to and picking up objects, after having learned a \emph{teacher's} language \cite{yu2017deep}. An RNN-based language module is connected to a CNN-based perception module. These two modules were then connected to an action-selection module and a recognition module that learns the teacher's language in a question answering process.

\subsection{Racing Games}
There are generally two paradigms for vision-based autonomous driving highlighted in Chen at al.~\cite{chen2015deepdriving}; (1) end-to-end systems that learn to map images to actions directly (behavior reflex), and (2) systems that parse the sensor data to make informed decisions (mediated perception). An approach that falls in between these paradigms is \emph{direct perception} where a CNN learns to map from images to meaningful affordance indicators, such as the car angle and distance to lane markings, from which a simple controller can make decisions \cite{chen2015deepdriving}. Direct perception was trained on recordings of 12 hours of human driving in TORCS and the trained system was able to drive in very diverse environments. Amazingly, the network was also able to generalize to real images. 

End-to-end reinforcement learning algorithms such as DQN cannot be directly applied to continuous environments such as racing games because the action space must be discrete and with relatively low dimensionality. Instead, policy gradient methods, such as actor-critic \cite{degris2012model} and Deterministic Policy Gradient (DPG) \cite{silver2014deterministic} can learn policies in high-dimensional and continuous action spaces. Deep DPG (DDPG) is a policy gradient method that implements both experience replay and a separate target network and was used to train a CNN end-to-end in TORCS from images \cite{lillicrap2015continuous}.

The aforementioned A3C methods have also been applied to the racing game TORCS using only pixels as input \cite{mnih2016asynchronous}. In those experiments, rewards were shaped as the agent's velocity on the track, and after 12 hours of training, A3C reached a score between roughly 75\% and 90\% of a human tester in tracks with and without opponent bots, respectively. 

While most approaches to training deep networks from high-dimensional input in video games are based on gradient descent, a notable exception is an approach by Koutn{\'\i}k et al.~\cite{koutnik2013evolving}, where Fourier-type coefficients were evolved that encoded a recurrent network with over 1 million weights. Here, evolution was able to find a high-performing controller for TORCS that only relied on high-dimensional visual input. 

\subsection{First-Person Shooters}
\label{sec:fps}

Kempka et al.~\cite{kempka2016vizdoom} demonstrated that a CNN with max-pooling and fully connected layers trained with DQN can achieve human-like behaviors in basic scenarios. In the Visual Doom AI Competition 2016\footnote{http://vizdoom.cs.put.edu.pl/competition-cig-2016}, a number of participants submitted pre-trained neural network-based agents that competed in a multi-player deathmatch setting. Both a \emph{limited} competition was held, in which bots competed in known levels, and a \emph{full} competition that included bots competing in unseen levels. 
The winner of the limited track used a CNN trained with A3C using reward shaping and curriculum learning \cite{wu2017training}. Reward shaping tackled the problem of sparse and delayed rewards, giving artificial positive rewards for picking up items and negative rewards for using ammunition and losing health. Curriculum learning attempts to speed up learning by training on a set of progressively harder environments \cite{bengio2009curriculum}. The second-place entry in the limited track used a modified DRQN network architecture with an additional stream of fully connected layers to learn supervised auxiliary tasks such as enemy detection, with the purpose of speeding up the training of the convolutional layers \cite{lample2016playing}. Position inference and object mapping from pixels and depth-buffers using Simultaneous Localization and Mapping (SLAM) also improve DQN in Doom \cite{bhatti2016playing}.


The winner of the full deathmatch competition implemented a \emph{Direct Future Prediction} (DFP) approach that was shown to outperform DQN and A3C
\cite{dosovitskiy2016learning}. The architecture used in DFP has three streams: one for the screen pixels, one for lower-dimensional measurements describing the agent's current state, and one for describing the agent's goal, which is a linear combination of prioritized measurements. DFP collects experiences in a memory and is trained with supervised learning techniques to predict the future measurements based on the current state, goal and selected action. During training, actions are selected that yield the best-predicted outcome, based on the current goal. This method can be trained on various goals and generalizes to unseen goals at test time.  

Navigation in 3D environments is one of the important skills required for FPS games and has been studied extensively. A CNN+LSTM network was trained with A3C extended with additional outputs predicting the pixel depths and loop closure, showing significant improvements \cite{mirowski2016learning}. 

The \emph{UNREAL} algorithm, based on A3C, implements an auxiliary task that trains the network to predict the immediate subsequent future reward from a sequence of consecutive observations. UNREAL was tested on fruit gathering and exploration tasks in OpenArena
and achieved a mean human-normalized score of 87\%, where A3C only achieved 53\% \cite{jaderberg2016reinforcement}.

The ability to transfer knowledge to new environments can reduce the learning time and can in some cases be crucial for some challenging tasks. Transfer learning can be achieved by pre-training a network in similar environments with simpler tasks or by using random textures during training \cite{chaplottransfer}. The \emph{Distill and Transfer Learning} (Distral) method trains several worker policies (one for each task) concurrently and shares a distilled policy ~\cite{teh2017distral}. The worker policies are regularized to stay close to the shared policy which will be the centroid of the worker policies. Distral was applied to DeepMind Lab.


The \emph{Intrinsic Curiosity Module} (ICM), consisting of several neural networks, computes an intrinsic reward each time step based on the agent's inability to predict the outcome of taking actions. It was shown to learn to navigate in complex Doom and Super Mario levels only relying on intrinsic rewards \cite{pathakICMl17curiosity}.



\subsection{Open-World Games}
The \emph{Hierarchical Deep Reinforcement Learning Network} (H-DRLN) architecture implements a lifelong learning framework, which is shown  to be able to transfer knowledge between simple tasks in Minecraft such as navigation, item collection, and placement tasks \cite{tessler2016deep}. H-DRLN uses a variation of policy distillation \cite{rusu2015policy} to retain and encapsulate learned knowledge into a single network. 

\emph{Neural Turing Machines} (NTMs) are fully differentiable neural networks coupled with an external memory resource, which can learn to solve simple algorithmic problems such as copying and sorting \cite{graves2014neural}. Two memory-based variations, inspired by NTM, called \emph{Recurrent Memory Q-Network} (RMQN) and \emph{Feedback Recurrent Memory Q-Network} (FRMQN) were able to solve complex navigation tasks that require memory and active perception \cite{oh2016control}. 

The Teacher-Student Curriculum Learning (TSCL) framework incorporates a teacher that prioritizes tasks wherein the student's performance is either increasing (learning) or decreasing (forgetting) \cite{matiisen2017teacher}. TSCL enabled a policy gradient learning method to solve mazes that were otherwise not possible with a uniform sampling of subtasks. 

\subsection{Real-Time Strategy Games}

The previous sections described methods that learn to play games \emph{end-to-end}, i.e.\ a neural network is trained to map states directly to actions. Real-Time Strategy (RTS) games, however, offer much more complex environments, in which players have to control multiple agents simultaneously in real-time on a partially observable map. Additionally, RTS games have no in-game scoring and thus the reward is determined by who wins the game. For these reasons, learning to play RTS games end-to-end may be infeasible for the foreseeable future and instead, sub-problems have been studied so far. 

For the simplistic RTS platform $\mu$RTS a CNN was trained as a state evaluator using supervised learning on a generated data set and used in combination with Monte Carlo Tree Search \cite{stanescu2016evaluating, barriga2017combining}. This approach performed significantly better than previous evaluation methods. 

StarCraft has been a popular game platform for AI research, but so far only with a few deep learning approaches. Deep learning methods for StarCraft have focused on micromanagement (unit control) or build-order planning and has ignored other aspects of the game. The problem of delayed rewards in StarCraft can be circumvented in combat scenarios; here rewards can be shaped as the difference between damage inflicted and damage incurred~\cite{usunier2016episodic, foerster2017stabilising, peng2017multiagent, foerster2017counterfactual}. States and actions are often described locally relative to units, which is extracted from the game engine. If agents are trained individually it is difficult to know which agents contributed to the global reward \cite{chang2003all}, a problem known as the \emph{multi-agent credit assignment problem}. One approach is to train a generic network, which controls each unit separately and search in policy space using Zero-Order optimization based on the reward accrued in each episode~\cite{usunier2016episodic}. This strategy was able to learn successful policies for armies of up to 15 units. 

\emph{Independent Q-learning} (IQL) simplifies the multi-agent RL problem by controlling units individually while treating other agents as if they were part of the environment \cite{tan1993multi}. This enables Q-learning to scale well to a large number of agents. However, when combining IQL with recent techniques such as experience replay, agents tend to optimize their policies based on experiences with obsolete policies. This problem is overcome by applying \emph{fingerprints} to experiences and by applying an importance-weighted loss function that naturally decays obsolete data, which has shown improvements for some small combat scenarios \cite{foerster2017stabilising}.

The \emph{Multiagent Bidirectionally-Coordinated Network} (BiCNet) implements a vectorized actor-critic framework based on a bi-directional RNN, with one dimension for every agent, and outputs a sequence of actions \cite{peng2017multiagent}. This network architecture is unique to the other approaches as it can handle an arbitrary number of units of different types.

\emph{Counterfactual multi-agent} (COMA) policy gradients is an actor-critic method with a centralized critic and decentralized actors that address the multi-agent credit assignment problem with a counterfactual baseline computed by the critic network \cite{foerster2017counterfactual}. COMA achieves state-of-the-art results, for decentralized methods, in small combat scenarios with up to ten units on each side.

Deep learning has also been applied to build-order planning in StarCraft using macro-based supervised learning approach to imitate human strategies \cite{justesen2017learning}. The trained network was integrated as a module used in an existing bot capable of playing the full game with an other-wise hand-crafted behavior. Another macro-based approach, here using RL instead of SL, called Convolutional Neural Network Fitted Q-Learning (CNNFQ), was trained with Double DQN for build-order planning in StarCraft II and was able to win against medium-level scripted bots on small maps \cite{tang2018reinforcement}. A macro action-based reinforcement learning method that uses Proximal Policy Optimization (PPO) for build order planning and high-level attack planning was able to outperform the built-in bot in StarCraft II at level 10 \cite{sun2018tstarbots}. This is particularly impressive as the level 10 bot cheats by having full vision of the map and faster resource harvesting. The results were obtained using 1920 parallel actors on 3840 CPUs across 80 machines and only for one matchup on one map. This system won a few games against Platinum-level human players but lost all games against Diamond-level players. The authors report that the learned policy "lacks strategy diversity in order to consistently beat human players" \cite{sun2018tstarbots}.

\subsection{Team Sports Games}

Deep Deterministic Policy Gradients (DDPG) was applied to RoboCup 2D Half-Field-Offense (HFO) \cite{hausknecht2015deep}. The actor network used two output streams, one for the selection of discrete action types (dash, turn, tackle, and kick) and one for each action type's 1-2 continuously-valued parameters (power and direction). The \emph{Inverting Gradients} bounding approach downscales the gradients as the output approaches its boundaries and inverts the gradients if the parameter exceeds them. This approach outperformed both SARSA and the best agent in the 2012 RoboCup. DDPG was also applied to HFO by mixing on-policy updates with 1-step Q-Learning updates \cite{hausknecht2016policy} and outperformed a hand-coded agent with expert knowledge with one player on each team.


\subsection{Physics Games}
As video games are usually a reflection or simplification of the real world, it can be fruitful to learn an intuition about the physical laws in an environment. A predictive neural network using an object-centered approach (also called fixations) learned to run simulations of a billiards game after being trained on random interactions \cite{fragkiadaki2015learning}. This predictive model could then be used for planning actions in the game. 

A similar predictive approach was tested in a 3D game-like environment, using the Unreal Engine, where ResNet-34 \cite{he2016deep} (a deep residual network used for image classification) was extended and trained to predict the visual outcome of blocks that were stacked such that they would usually fall \cite{lerer2016learning}. Residual networks implement \emph{shortcut connections} that skip layers, which can improve learning in very deep networks.

\subsection{Text Adventure Games}

Text adventure games, in which both states and actions are presented as text only, are a special video game genre. A network architecture called LSTM-DQN \cite{narasimhan2015language} was designed specifically to play these games and is implemented using LSTM networks that convert text from the world state into a vector representation, which estimates Q-values for all possible state-action pairs. LSTM-DQN was able to complete between 96\% and 100\% of the quests on average in two different text adventure games.

To be able to improve on these results, researchers have moved toward learning language models and word embeddings to augment the neural network. An approach that combines reinforcement learning with explicit language understanding is Deep Reinforcement Relevance Net (DRRN) \cite{he2016drrn}. This approach has two networks that learn word embeddings. One embeds the state description, the other embeds the action description. Relevance between the two embedding vectors is calculated with an interaction function such as the inner product of the vectors or a bilinear operation. The Relevance is then used as the Q-Value and the whole process is trained end-to-end with Deep Q-Learning. This approach allows the network to generalize to phrases not seen during training which is an improvement for very large text games. The approach was tested on the text games Saving John and Machine of Death, both choice-based games.

Taking language modeling further, Fulda et. al. explicitly modeled language affordances to assist in action selection \cite{fulda2017can}. A word embedding is first learned from a Wikipedia Corpus via unsupervised learning \cite{mikolov2013efficient} and this embedding is then used to calculate analogies such as \emph{song is to sing} as \emph{bike is to $x$}, where $x$ can then be calculated in the embedding space \cite{mikolov2013efficient}. The authors build a dictionary of verbs, noun pairs, and another one of object manipulation pairs. Using the learned affordances, the model can suggest a small set of actions for a state description. Policies were learned with Q-Learning and tested on 50 Z-Machine games.

The Golovin Agent focuses exclusively on language models \cite{kostka2017text} that are pre-trained from a corpus of books in the fantasy genre. Using word embeddings, the agent can replace synonyms with known words. Golovin is built of five command generators: General, Movement, Battle, Gather, and Inventory. These are generated by analyzing the state description, using the language models to calculate and sample from a number of features for each command. Golovin uses no reinforcement learning and scores comparable to the affordance method.

Most recently, Zahavy et. al. proposed another DQN method \cite{zahavy2018learn}. This method uses a type of attention mechanism called Action Elimination Network (AEN). In parser-based games, the actions space is very large. The AEN learns, while playing, to predict which actions that will have no effect for a given state description. The AEN is then used to eliminate most of the available actions for a given state and after which the remaining actions are evaluated with the Q-network. The whole process is trained end-to-end and achieves similar performance to DQN with a manually constrained actions space. Despite the progress made for text adventure games, current techniques are still far from matching human performance.

Outside of text adventure games, natural language processing has been used for other text-based games as well. To facilitate communication, a deep distributed recurrent Q-network (DDRQN) architecture was used to train several agents to learn a communication protocol to solve the multi-agent \emph{Hats} and \emph{Switch} riddles \cite{foerster2016learning}. One of the novel modifications in DDRQN is that agents use shared network weights that are conditioned on their unique ID, which enables faster learning while retaining diversity between agents.






\section{Historical Overview of Deep Learning in Games}
\label{sec:history}
The previous section discussed deep learning methods in games according to the game type. This section instead looks at the development of these methods in terms of how they influenced each other, giving a historical overview of the deep learning methods that are reviewed in the previous section. Many of these methods are inspired from or directly build upon previous methods, while some are applied to different game genres and others are tailored to specific types of games.

\begin{figure*}
\begin{center}
  \includegraphics[width=.8\textwidth]{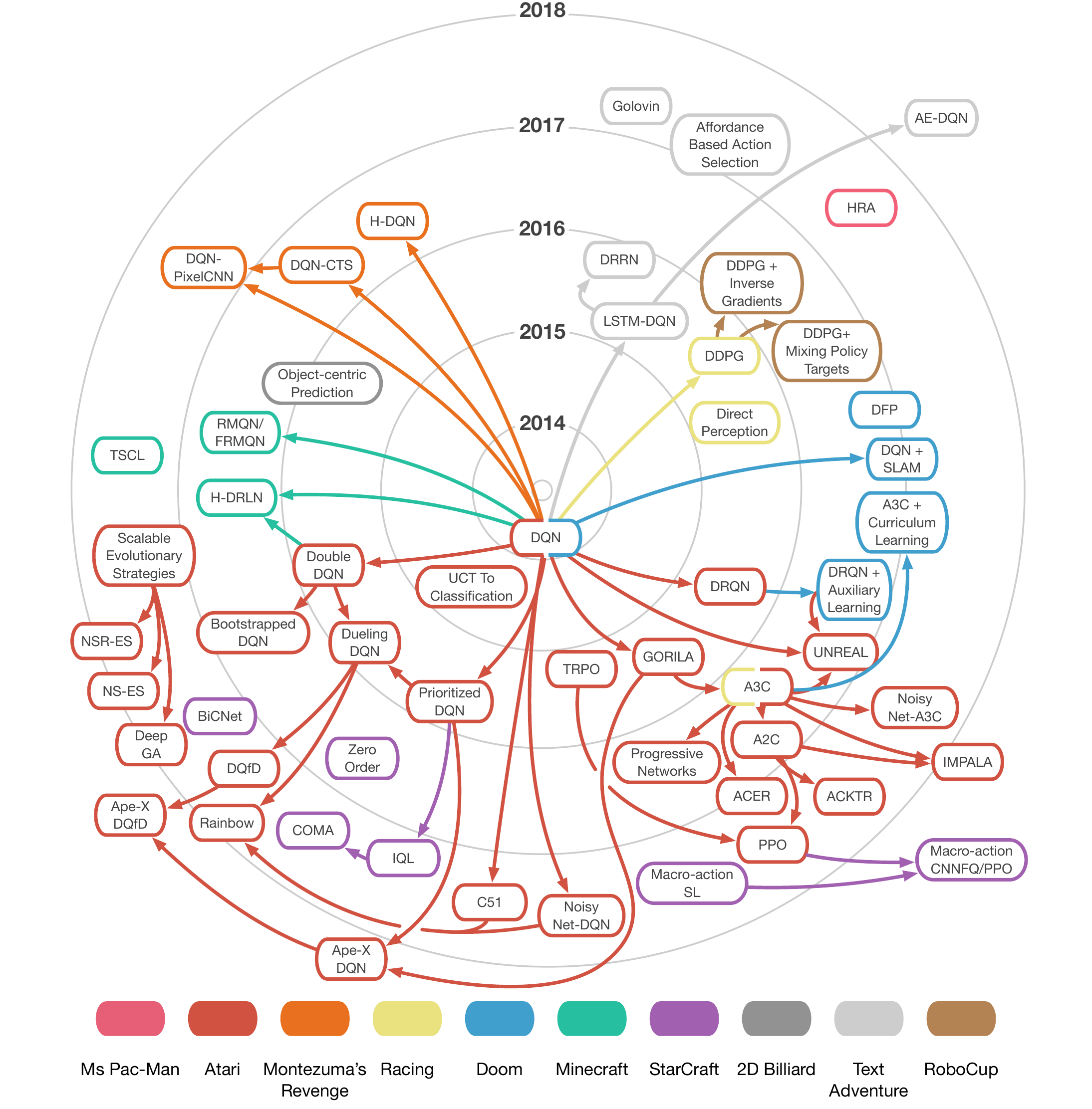} 
  \caption{
   Influence diagram of the deep learning techniques discussed in this paper. Each node is an algorithm while the color represents the game benchmark. The distance from the center represents the date that the original paper was published on arXiv. The arrows represent how techniques are related. Each node points to all other nodes that used or modified that technique. Arrows pointing to a particular algorithm show which algorithms influenced its design. Influences are not transitive: if algorithm a influenced b and b influenced c, a did not necessarily influence c.} 
  \label{fig:history}
  \vspace{-2.0em}
\end{center}
\end{figure*}

Figure~\ref{fig:history} shows an influence diagram with the reviewed methods and their relations to earlier methods (the current section can be read as a long caption to that figure). Each method in the diagram is colored to show the game benchmark. DQN \cite{mnih2013playing} was very influential as an algorithm that uses gradient-based deep learning for pixel-based video game playing and was originally applied to the Atari benchmark. Note that earlier approaches exist but with less success such as~\cite{parker2012neurovisual}, and successful gradient-free methods~\cite{risi2015neuroevolution}. Double DQN \cite{van2016deep} and Dueling DQN \cite{wang2015dueling} are early extensions that use multiple networks to improve estimations. DRQN \cite{hausknecht2015deep} uses a recurrent neural network as the Q network. Prioritized DQN \cite{schaul2015prioritized} is another early extension and it adds improved experience replay sampling. Bootstrapped DQN \cite{osband2016deep} builds off of Double DQN with a different improved sampling strategy. Further DQN enhancements used for Atari include: the C51 algorithm~\cite{bellemare2017distributional}, which is based on DQN but changes the Q function; Noisy-Nets which make the networks stochastic to aid with exploration \cite{fortunato2018noisy}; DQfD which also learns from examples \cite{hester2017deep}; and Rainbow, which combines many of these state-of-the-art techniques together \cite{hessel2018rainbow}.

Gorila was the first asynchronous method based on DQN \cite{nair2015massively} and was followed by A3C \cite{mnih2016asynchronous} which uses multiple asynchronous agents for an actor-critic approach. This was further extended at the end of 2016 with UNREAL \cite{jaderberg2016reinforcement}, which incorporates work done with auxiliary learning to handle sparse feedback environments. Since then there has been a lot of additional extensions on A3C \cite{wu2017scalable}, \cite{wang2016sample}, \cite{rusu2016progressive}, \cite{fortunato2018noisy}. IMPALA has taken it further with focusing on a single trained agent that can play all of the Atari games \cite{espeholt2018impala}. In 2018, the move toward large scale distributed learning has continued and advanced with Ape-X \cite{horgan2018distributed, pohlen2018observe}.

Evolutionary techniques are also seeing a Renaissance for video games. First Salimans et. al. showed that Evolution Strategies could compete with deep RL \cite{salimans2017evolution}. Then two more papers came out of Uber AI: one showing that derivative-free evolutionary algorithms can compete with deep RL \cite{such2017deep}, and an extension to ES \cite{conti2018improving}. These benefit from easy parallelization and possibly have some advantage in exploration.

Another approach used on Atari around the time that DQN was introduced is Trust Region Policy Optimization \cite{kulkarni2016hierarchical}. This updates a surrogate objective that is updated from the environment. Later in 2017, Proximal Policy Optimization was introduced as a more robust, simpler surrogate optimization scheme that also draws from innovations in A3C \cite{schulman2017proximal}. Some extensions are specifically for Montezuma's revenge, which is a game within the ALE benchmark, but it is particularly difficult due to sparse rewards and hidden information. The algorithms that do best on Montezuma do so by extending DQN with intrinsic motivation~\cite{bellemare2016unifying} and hierarchical learning \cite{kulkarni2016hierarchical}. Ms. Pack-Man was also singled out from Atari, where the reward function was learned in separate parts to make the agent more robust to new environments~\cite{van2017hybrid}. 

Doom is another benchmark that is new as of 2016. Most of the work for this game has been extending methods designed for Atari to handle richer data. A3C + Curriculum Learning \cite{wu2017training} proposes using curriculum learning with A3C. DRQN + Auxiliary Learning \cite{lample2016playing} extends DRQN by adding additional rewards during training. DQN + SLAM \cite{bhatti2016playing} combines techniques for mapping unknown environments with DQN. 

DFP \cite{dosovitskiy2016learning} is the only approach that is not extending an Atari technique. Like UCT To Classification \cite{guo2014deep} for Atari, Object-centric Prediction \cite{fragkiadaki2015learning} for Billiard, and Direct Perception \cite{chen2015deepdriving} for Racing, DFP uses supervised learning to learn about the game. All of these, except UCT To Classification, learn to directly predict some future state of the game and make a prediction from this information. None of these works, all from different years, refer to each other. Besides Direct Perception, the only unique work for racing is Deep DPG \cite{lillicrap2015continuous}, which extends DQN for continuous controls. This technique has been extended for RoboCup Soccer \cite{hausknecht2015deepparam} \cite{hausknecht2016policy}.

Work on StarCraft micro-management (unit control) is based on Q-learning started in late 2016. IQL \cite{foerster2017stabilising} extends DQN Prioritized DQN by treating all other agents as part of the environment. COMA \cite{foerster2017counterfactual} extends IQL by calculating counterfactual rewards, the marginal contribution each agent added. biCNet \cite{peng2017multiagent} and Zero Order Optimization~\cite{usunier2016episodic}, are reinforcement learning based but are not derived from DQN. Another popular approach is hierarchical learning. In 2017 it was tried with replay data \cite{justesen2017learning} and in 2018 state of the art results were achieved by using it with two different RL methods \cite{sun2018tstarbots,tang2018reinforcement}.

Some work published in 2016 extends DQN to play Minecraft \cite{tessler2016deep}. At around the same time, techniques were developed to make DQN context-aware and modular to handle the large state space \cite{oh2016control}. Recently, curriculum learning has been applied to Minecraft as well \cite{matiisen2017teacher}.

DQN was applied to text adventure games in 2015 \cite{narasimhan2015language}. Soon after, it was modified to have a language-specific architecture and use the state-action pair relevance as the Q value \cite{he2016drrn}. Most of the work on these games has been focused on explicit language modeling. Golovin Agent and Affordance Based Action Selection both use neural networks to learn language models which provide the actions for the agents to play \cite{fulda2017can,kostka2017text}. Recently, in 2018, DQN was used again paired with an Action Elimination Network \cite{zahavy2018learn}.

Combining extensions from previous algorithms have proven to be a promising direction for deep learning applied to video games, with Atari being the most popular benchmark for RL. Another clear trend, which is apparent in Table \ref{tab:overview}, is the focus on parallelization: distributing the work among multiple CPUs and GPUs. Parallelization is most common with actor-critic methods, such as A2C and A3C, and evolutionary approaches, such as Deep GA \cite{such2017deep} and Evolution Strategies \cite{salimans2017evolution,conti2018improving}. Hierarchical reinforcement learning, intrinsic motivation, and transfer learning are promising new directions to explore to master currently unsolved problems in video game playing.

\section{Open Challenges}
\label{sec:open}
While deep learning has shown remarkable results in video game playing, a multitude of important open challenges remain, which we review here. Indeed, looking back at the current state of research from a decade or two in the future, it is likely that we will see the current research as early steps in a broad and important research field. This section is divided into four broad categories (agent model properties, game industry, learning models of games, and computational resources) with different game-playing challenges that remain open for deep learning techniques. We mention a few potential approaches for some of the challenges while the best way forward for others is currently not clear.

\subsection{Agent Model Properties}
\subsubsection{General Video Game Playing}

Being able to solve a single problem does not make you intelligent; nobody would say that Deep Blue or AlphaGo \cite{silver2016mastering} possess general intelligence, as they cannot even play Checkers (without re-training), much less make coffee or tie their shoelaces. To learn generally intelligent behavior, you need to train on not just a single task, but many different tasks~\cite{legg2007universal}. Video games have been suggested as ideal environments for learning general intelligence, partly because there are so many video games that share common interface and reward conventions~\cite{schaul2011measuring}. Yet, the vast majority of work on deep learning in video games focuses on learning to play a single game or even performing a single task in a single game.

While deep RL-based approaches can learn to play a variety of different Atari games, it is still a significant challenge  to develop algorithms that can learn to play any kind of game (e.g.\ Atari games, DOOM, and StarCraft). Current approaches still require significant effort to design the network architecture and reward function to a specific type of game. 

Progress on the problem of playing multiple games includes progressive neural networks \cite{rusu2016progressive}, which allow new games to be learned (without forgetting previously learned ones) and solved quicker by exploiting previously learned features through lateral connections. However, they require a separate network for each task. Elastic weight consolidation \cite{kirkpatrick2017overcoming} can learn multiple Atari games sequentially and avoids catastrophic forgetting by protecting weights from being modified that are important for previously learned games. In PathNet an evolutionary algorithm is used to select which parts of a neural network are used for learning new tasks, demonstrating some transfer learning performance on ALE games~\cite{fernando2017pathnet}.

In the future it will be important to extend these methods to learn to play multiple games, even if those games are very different --- most current approaches focus on different (known) games in the ALE framework. One suitable avenue for this kind of research is the new Learning Track of the GVGAI competition~\cite{kunanusont2017general,torrado2018deep}. GVGAI has a potentially unlimited set of games, unlike ALE. Recent work in GVGAI showed that model-free deep RL overfitted not just to the individual game, but even to the individual level; this was countered by continuously generating new levels during training~\cite{justesen2018illuminating}. 

It is possible that significant advances on the multi-game problem will come from outside deep learning. In particular, the recent Tangled Graph representation, a form of genetic programming, has shown promise in this task~\cite{kelly2017multi}. The recent IMPALA algorithm tries to tackle multi-game learning through massive scaling, with somewhat promising results~\cite{espeholt2018impala}.


\subsubsection{Overcoming sparse, delayed, or deceptive Rewards}
Games such as Montezuma's Revenge that are characterized by sparse rewards still pose a challenge for most Deep RL approaches. While recent advances that combine DQN with intrinsic motivation \cite{bellemare2016unifying} or expert demonstrations \cite{hester2017deep, pohlen2018observe} can help, games with sparse rewards are still a challenge for current deep RL methods. There is a long history of research in intrinsically motivated reinforcement learning~\cite{chentanez2005intrinsically,schmidhuber2010formal} as well as hierarchical reinforcement learning which might be useful here~\cite{barto2003recent,wiering1997hq}. The Project Malmo environment, based on Minecraft, provides an excellent venue for creating tasks with very sparse rewards where agents need to set their own goals. Derivative-free and gradient-free methods, such as  evolution strategies and genetic algorithms, explore the parameter space by sampling locally and are promising for these games, especially when combined with novelty search as in \cite{conti2018improving, such2017deep}.

\subsubsection{Learning with multiple agents}
Current deep RL approaches are mostly concerned with training a single agent. A few exceptions exist where multiple agents have to cooperate \cite{leibo2017multi, foerster2017stabilising, usunier2016episodic, peng2017multiagent, foerster2017counterfactual}, but it remains an open challenge how these can scale to more agents in various situations. In many current video games such as \emph{StarCraft} or \emph{GTA V}, many agents interact with each other and the player. To scale multi-agent learning in video games to the same level of performance as current single agent approaches will likely require new methods that can effectively train multiple agents at the same time.

\subsubsection{Lifetime adaptation}
While NPCs can be trained to play a variety of games well (see Section~\ref{sec:game_playing}), current machine learning techniques still struggle when it comes to agents that should be able to adapt during their lifetime, i.e.\ while the game is being played. For example, a human player can quickly change its behavior when realizing that the player is always ambushed at the same position in an FPS map. However, most current DL techniques would require expensive re-training to adapt to these situations and other unforeseen situations that they have not encountered during training. The amount of data provided by the real-time behavior of a single human is nowhere near that required by common deep learning methods. This challenge is related to the wider problem of few-shot learning, transfer learning and general video game playing. Solving it will be important to create more believable and human-like NPCs.

\subsubsection{Human-like game playing}
Lifetime learning is just one of the differences that current NPCs lack in comparison to human players. Most approaches are concerned with creating agents that play a particular game as well as possible, often only taking into account the score reached. However, if humans are expected to play against or cooperate with AI-based bots in video games, other factors come into play. Instead of creating a bot that plays perfectly, in this context it becomes more important that the bot is believable and is fun to play against, with similar idiosyncrasies we expect from a human player. 

Human-like game playing is an active area of research with two different competitions focused on human-like behavior namely the \emph{2k BotPrize}~\cite{hingston2010new,hingston2012believable} and the Turing Test track of the \emph{Mario AI Championship}~\cite{shaker2013turing}. Most entries in these competitions are based on various non-neural network techniques while some used evolutionary training of deep neural networks to generate human-like behavior~\cite{schrum2011ut,ortega2013imitating}.


\subsubsection{Adjustable performance levels}

Almost all current research on DL for game playing aims at creating agents that can play the game as well as possible, maybe even ``beating'' it. However, for purposes of both game testing, creating tutorials, and demonstrating games---in all those places where it would be important to have human-like game `play---it could be important to be able to create agents with a particular skill level. If your agent plays better than any human player, then it is not a good model of what a human would do in the game. At its most basic, this could entail training an agent that plays the game very well, and then find a way of decreasing the performance of that agent. However, it would be more useful to be able to adjust the performance level in a more fine-grained way, so as to for example separately control the reaction speed or long-term planning ability of an agent. Even more useful would be to be able to ban certain capacities of playstyles of a trained agent, so a to test whether for example a given level could be solved without certain actions or tactics.

One path to realizing this is the concept of \emph{procedural personas}, where the preferences of an agent are encoded as a set of utility weights~\cite{holmgaard2014generative}. However, this concept has not been implemented using deep learning, and it is still unclear how to realize the planning depth control in this context.

\subsubsection{Dealing with extremely large decision spaces}
Whereas the average branching factor hovers around 30 for Chess and 300 for Go, a game like StarCraft has a branching factor that is orders of magnitudes larger. While recent advances in evolutionary planning have allowed real-time and long-term planning in games with larger branching factors to \cite{justesen2016online,wang2016portfolio,justesen2017continual}, how we can scale Deep RL to such levels of complexity is an important open challenge. Learning heuristics with deep learning in these games to enhance search algorithms is also a promising direction.


\subsection{Game Industry}
\subsubsection{Adoption in the game industry}
Many of the recent advances in DL have been accelerated because of the increased interest by a variety of different companies such as Facebook, Google/Alphabet, Microsoft and Amazon, which heavily invest in its development. However, the game industry has not embraced these advances to the same extent. This sometimes surprises commentators outside of the game industry, as games are seen as making heavy use of AI techniques. However, the type of AI that is  most commonly used in the games industry focuses more on hand-authoring of expressive non-player character (NPC) behaviors rather than machine learning. An often-cited reason for the lack of adoption of neural networks (and similar methods) within this industry is that such methods are inherently difficult to control, which could result in unwanted NPC behaviors (e.g.\ an NPC could decide to kill a key actor that is relevant to the story). Additionally, training deep network models require a certain level of expertise and the pool of experts in this area is still limited. It is important to address these challenges to encourage a wide adoption in the game industry. 

Additionally, while most DL approaches focus exclusively on playing games as well as possible, this goal might not be the most important for the game industry~\cite{yannakakis2017artificial}. Here  the level of fun or engagement the player experiences while playing is a crucial component. One use of DL for game playing in the game production process is for game testing, where artificial agents test that levels are solvable or that the difficulty is appropriate. DL might see its most prominent use in the games industry not for playing games, but for generating game content~\cite{shaker2016procedural} based on training on existing content ~\cite{summerville2018procedural,volz2018evolving}, or for modeling player experience~\cite{yannakakis2013player}. 



Within the game industry, several of the large development and technology companies, including Electronic Arts, Ubisoft and Unity have recently started in-house research arms focusing partly on deep learning. It remains to be seen whether these techniques will also be embraced by the development arms of these companies or their customers. 

\subsubsection{Interactive tools for game development}
Related to the previous challenge, there is currently a lack of tools for designers to easily train NPC behaviors. While many open-source tools to training deep networks exist now, most of them require a significant level of expertise. A tool that allows designers to easily specify desired NPC behaviors (and undesired ones) while assuring a certain level of control over the final trained outcomes would greatly accelerate the uptake of these new methods in the game industry. 

Learning from human preferences is one promising direction in this area. This approach has been extensively studied in the context of neuroevolution \cite{risi2015neuroevolution}, and also in the context of video games,  allowing non-expert users to breed behaviors for Super Mario \cite{sorensen2016breeding}. Recently a similar preference-based approach was applied to deep RL method \cite{christiano2017deep}, allowing agents to learn Atari games based on a combination of human preference learning and deep RL. Recently, the game company King published results using imitation learning to learn policies for play-testing of Candy Crush levels, showing a promising direction for new design-tools \cite{gudmundsson2017human}.

\subsubsection{Creating new types of video games}
DL could potentially offer a way to create completely new games. Most of today's game designs stem from a time when no advanced AI methods were available or the hardware too limited to utilize them, meaning that games have been designed to not need AI. Designing new games \emph{around} AI can help to break out of these limitations. While evolutionary algorithms and neuroevolution in particular \cite{risi2015neuroevolution} have allowed the creation of completely new types of games, DL based on gradient descent has not been explored in this context. Neuroevolution is a core mechanic in games such as NERO~\cite{stanley2005real}, Galactic Arms Race~\cite{hastings2009automatic}, Petalz~\cite{risi2012combining} and EvoCommander~\cite{jallov2017evocommander}. One challenge with gradient-based optimization is that the structures are often limited to having mathematical smoothness (i.e.\ differentiability), making it challenging to create interesting and unexpected outputs.

\subsection{Learning models of games}

Much work on deep learning for game-playing takes a model-free end-to-end learning approach, where a neural network is trained to produce actions given state observations as input. However, it is well known that a good and fast forward model makes game-playing much easier, as it makes it possible to use planning methods based on tree search or evolution~\cite{yannakakis2017artificial}. Therefore, an important open challenge in this field is to develop methods that can learn a forward model of the game, making it possible to reason about its dynamics.

The hope is that approaches that learn the rules of the game can generalize better to different game variations and show more robust learning. Promising work in this area  includes the approach by Guzdial et al.~\cite{Guzdial2017} that learns a simple game engine of Super Mario Bros. from gameplay data. Kansky et al.~\cite{kansky2017schema} introduce the idea of Schema Networks that follow an object-oriented approach and are trained to predict future object attributes and rewards based on the current attributes and actions. A trained schema network thus provides a probabilistic
model that can be used for planning and is able to perform zero-shot transfer to variations of Breakout similar to those used in training. 

\subsection{Computational Resources}
With more advanced computational models and a larger number of agents in  open-worlds, computational speed becomes a concern. Methods that aim to make the networks computationally more efficient by either compressing networks \cite{iandola2016squeezenet} or pruning networks after training \cite{hassibi1993second,guo2016dynamic} could be useful. Of course, improvements in processing power in general or for neural networks specifically will also be important. Currently, it is not feasible to train networks in real-time to adapt to changes in the game or to fit players' playing styles, something which could be useful in the design process.






\section{Conclusion}
\label{sec:conclusion}
This paper reviewed deep learning methods applied to game playing in video games of various genres including; arcade, racing, first-person shooters, open-world, real-time strategy, team sports, physics, and text adventure games. Most of the reviewed work is within end-to-end model-free deep reinforcement learning, where a convolutional neural network learns to play directly from raw pixels by interacting with the game. Recent work demonstrates that derivative-free evolution strategies and genetic algorithms are competitive alternatives. Some of the reviewed work apply supervised learning to imitate behaviors from game logs, while others are based on methods that learn a model of the environment. For simple games, such as most arcade games, the reviewed methods can achieve above human-level performance, while there are many open challenges in more complex games.

\section*{Acknowledgements}
We thank the numerous colleagues who took the time to comment on drafts of this article, including Chen Tessler,
Diego P\'erez-Li\'ebana,
Ethan Caballero,
Hal Daum\'e III,
Jonas Busk,
Kai Arulkumaran,
Malcolm Heywood,
Marc G. Bellemare,
Marc-Philippe Huget,
Mike Preuss,
Nando de Freitas,
Nicolas A. Barriga,
Olivier Delalleau,
Peter Stone,
Santiago Onta\~n\'on,
Tambet Matiisen,
Yong Fu, and
Yuqing Hou.

\bibliographystyle{abbrv}
\bibliography{references}

\end{document}